\documentclass[lettersize,journal]{IEEEtran}
\DeclareUnicodeCharacter{FF1A}{:}
\usepackage{amsmath,amsfonts}
\usepackage{algorithm}
\usepackage{array}
\usepackage[caption=false,font=normalsize,labelfont=sf,textfont=sf]{subfig}
\usepackage{textcomp}
\usepackage{stfloats}
\usepackage{verbatim}
\usepackage{graphicx}
\usepackage{cite}
\usepackage{algpseudocode}
\usepackage{enumitem}
\usepackage{booktabs}
\usepackage{makecell}
\usepackage{siunitx} 
\usepackage{tikz}
\definecolor{citecolor}{RGB}{119,185,0} 
\usepackage[pagebackref=false,breaklinks=true,letterpaper=true,colorlinks,citecolor=citecolor,bookmarks=false]{hyperref}
\usepackage{multirow}
\newcommand{\PinkComment}[1]{\State \textcolor{blue}{\# #1}}

\def\eg{\emph{e.g.}} 
\def\ie{\emph{i.e.}}

\def\etal{\emph{et al.}}

\begin{document}

\title{
CAMeL: Cross-modality Adaptive Meta-Learning 
\\for Text-based Person Retrieval}

\author{Hang Yu, ~\IEEEmembership{Member, ~IEEE}, Jiahao Wen, and Zhedong Zheng, ~\IEEEmembership{Member, ~IEEE}
\thanks{Manuscript received xxx; accepted xxx. date of publication xxx; date of current version xxx. (Corresponding author: Zhedong Zheng. )}
\thanks{Hang Yu and Jiahao Wen are with the School of Computer Engineer and Science, Shanghai University, 
Shanghai 200444, China (e-mail: yuhang@shu.edu.cn; wenjh@shu.edu.cn). }
\thanks{Zhedong Zheng is with the Faculty of Science and Technology, and Institute of Collaborative Innovation, University of Macau, Macau 999078, China (e-mail: (zhedongzheng@um.edu.mo)}}

\markboth{Journal of \LaTeX\ Class Files,~Vol.~14, No.~8, August~2021}%
{Shell \MakeLowercase{\textit{et al.}}: A Sample Article Using IEEEtran.cls for IEEE Journals}


\maketitle
\begin{tikzpicture}[remember picture,overlay]
\node[anchor=north west,xshift=3.25cm,yshift=-1.75cm] at (current page.north west) 
{\includegraphics[width=0.75cm]{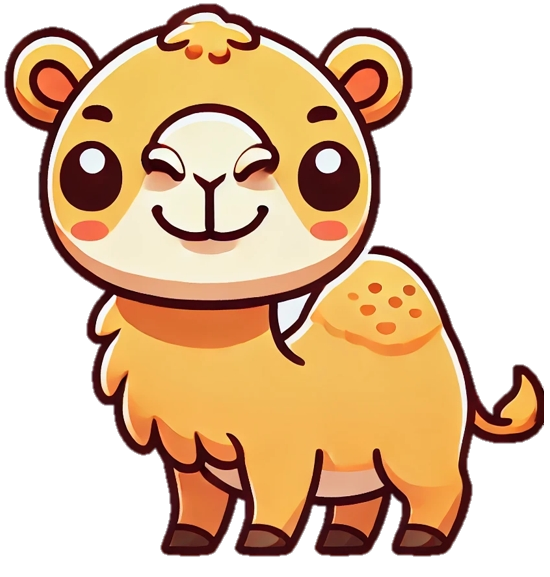}};
\end{tikzpicture}

\begin{abstract}
Text-based person retrieval aims to identify specific individuals within an image database using textual descriptions.
Due to the high cost of annotation and privacy protection, researchers resort to synthesized data for the paradigm of pretraining and fine-tuning. However, these generated data often exhibit domain biases in both images and textual annotations, which largely compromise the scalability of the pre-trained model. Therefore, we introduce a domain-agnostic pretraining framework based on Cross-modality Adaptive Meta-Learning (CAMeL) to enhance the model generalization capability during pretraining to facilitate the subsequent downstream tasks.  
In particular, we develop a series of tasks that reflect the diversity and complexity of real-world scenarios, and introduce a dynamic error sample memory unit to memorize the history for errors encountered within multiple tasks. 
To further ensure multi-task adaptation, we also adopt an adaptive dual-speed update strategy, balancing fast adaptation to new tasks and slow weight updates for historical tasks.
Albeit simple, our proposed model not only surpasses existing state-of-the-art methods on real-world benchmarks, including CUHK-PEDES, ICFG-PEDES, and RSTPReid, but also showcases robustness and scalability in handling biased synthetic images and noisy text annotations. Our code is available at https://github.com/Jahawn-Wen/CAMeL-reID.
\end{abstract}


\section{Introduction}
\IEEEPARstart{T}{ext}-based person retrieval tasks are closely linked with pedestrian re-identification~\cite{book,xiao2016end4} and text-image retrieval~\cite{jiang2023cross6, 10.1145/3543507.3583254}, aiming to identify specific individuals within an image database using textual queries~\cite{li2017person1, nguyen2024ag, cheng2024neighbor101, lu2025nighttime100, liu2024improving102, ye2024securereid103}. With the development of multimodal technologies and the intuitive nature of the text, the demand for matching images using natural language descriptions has been increasingly growing. 

\begin{figure}
    \centering
    \includegraphics[width=1\linewidth]{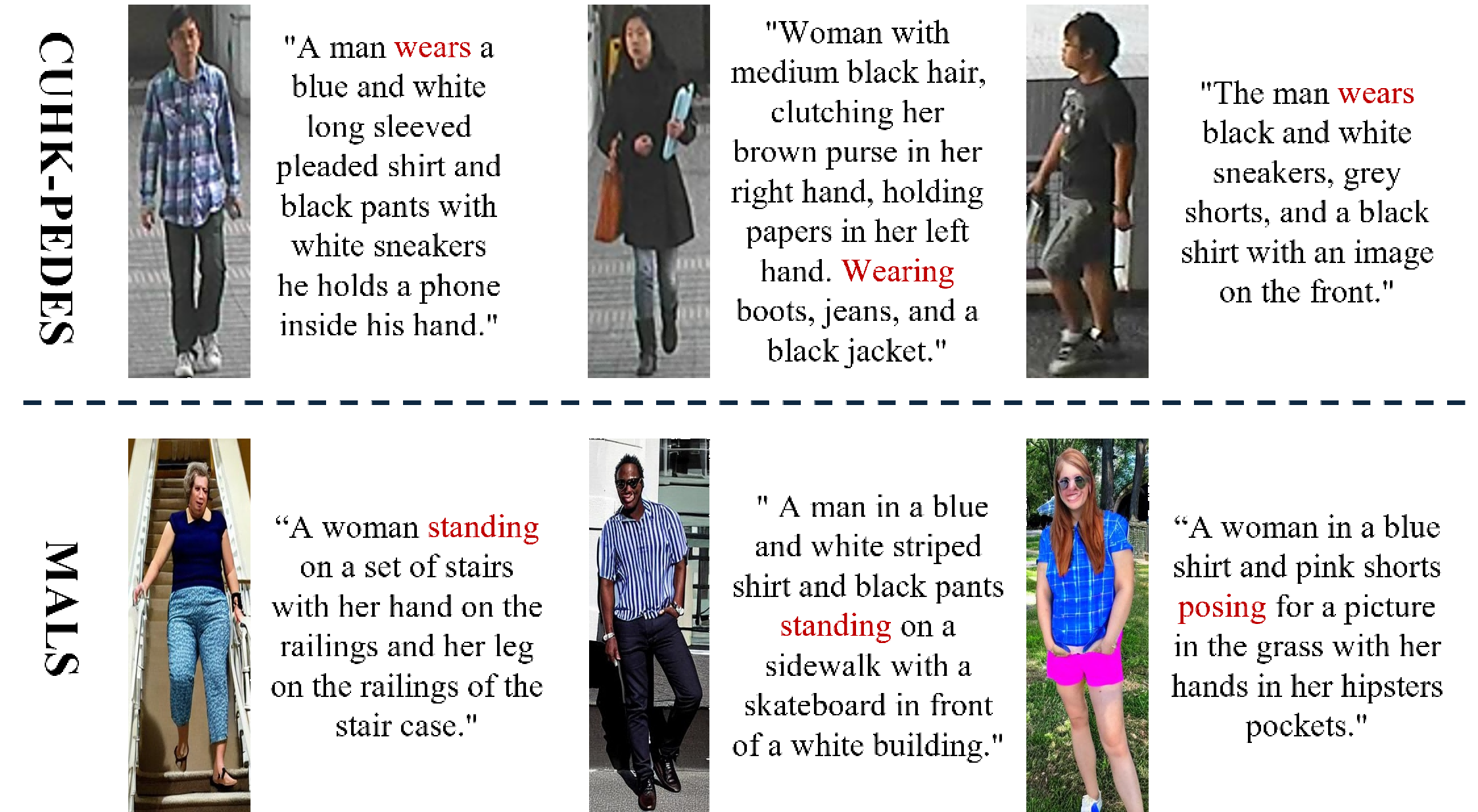}
    \caption{
    Domain biases are observed between the real-world dataset, CUHK-PEDES (top)~\cite{li2017person1}, and the synthetic dataset, MALS (bottom)~\cite{yang2023towards11}. The visual domain gap includes facial texture defects, resolution differences, and variations in illumination and color. Text annotations also exhibit bias, with MALS favoring gerunds such as ``standing" and ``posing," while CUHK-PEDES uses more specific verbs, \eg, ``wears."}
    \vspace{-.1in}
    \label{fig1}
\end{figure}

Compared to traditional image matching methods~\cite{li2018richly64}, utilizing natural language descriptions as a query not only simplifies the process of query design, but also enhances the flexibility and intuitiveness of the search. However, due to annotation difficulty and privacy issues, acquiring a large volume of real data for model training is often challenging~\cite{zheng2017unlabeled, bertocco2021unsupervised}. In an attempt to address the data scarcity, some researchers~\cite{yang2023towards11, chu2024towards, yao2023capenrich} have leveraged existing generative models, such as diffusion models~\cite{rombach2022high9, hertz2022prompt10}, to synthesize more training image-text pairs. Although involving generated data facilitates model training, the stylistic biases present in the generated images and textual descriptions often fail to capture the expected representations, thereby limiting the  generalization capability of the learned model (see Fig.~\ref{fig1}).

\textbf{Why there are visual domain gaps between real and synthetic data, such as discrepancies in illumination and color, facial texture defects, and resolution?} 
\textit{1) Challenges in Reproducing Real-World Illumination and Color Consistency.}
Real-world images are characterized by complex and diverse lighting conditions, including variations in light sources, shadows, and reflections. Additionally, color consistency is influenced by factors such as the color profile of camera, ambient lighting, and post-processing. Generative models often fail to fully capture these nuances, leading to discrepancies in illumination and color. Generated images typically display uniform or unrealistic lighting, and colors often fail to match the subtle variations and natural gradients found in real images.
This results in a visual domain gap, where synthetic images appear less realistic and more artificial.
For instance, synthetic pedestrian images show flat or uniform lighting, lacking the dynamic shadows and highlights that are typical in real photographs. 
Colors in the synthetic images also appear washed out or oversaturated, failing to replicate the natural skin tones and environmental colors present in real-world scenes.
\textit{2) Inadequate Modeling of High-Frequency Details and Textural Variations.} Generative models, such as GANs and diffusion models, often struggle to accurately capture and reproduce the high-frequency details and textural variations present in real-world images~\cite{karras2017progressive}. This is due to inherent model limitations and the complexity of natural textures. The result is that synthetic images usually exhibit artifacts such as blurring, smoothing, or unnatural patterns, which are particularly noticeable in fine-grained regions like skin and hair. These defects can manifest as inconsistencies in facial textures, leading to a clear difference between real and generated images.
For example, in some synthetic pedestrian images, the skin appear overly smooth or exhibits unnatural blemishes, while hair lacks the fine details and natural flow seen in real images.

\textbf{Similarly, we face the textual domain gap in the pedestrian descriptions. } 
To mitigate the image generation failures, such as missing attributes, we typically regenerate captions for synthesized images using the off-the-shelf captioning model. It also introduce the textual biases, which can be attributed to the training data distribution and the way the captioning model is trained. 
If the dataset used for training the captioning model contains a higher frequency of gerunds, the model will learn to generate captions that reflect this pattern. This is because the training objective is to minimize the loss function, which often leads it to reproduce the most common patterns in the training data. 
Consequently, the model overfits to the use of gerunds and fail to generalize well to the more varied and contextually rich verb usage found in real-world datasets.
For instance, if the training dataset frequently includes images with captions like ``A person standing in front of a building" or ``A woman posing for a photo," the captioning model will learn to prefer these gerund forms. 
In contrast, real-world datasets have more diverse and specific verb usages, such as ``A person wears a red jacket" or ``A woman smiles at the camera."
Besides, the usage of gerunds in captioning models often simplifies or generalizes actions, making them easier for the model to learn and generate.
In a real-world dataset, an image of a person wearing a hat and holding a book is captioned as ``A person wears a hat and holds a book." 
The verbs ``wears" and ``holds" provide specific and detailed information about the actions. 
In a synthetic dataset, the same image could be captioned as ``A person standing and holding a book," where ``standing" is a more general description of the state.

Considering the domain gap between the real-world data and generated data, we introduce a domain-agnostic pretraining framework for text-based person retrieval using Cross-modality Adaptive Meta-Learning (CAMeL). 
In particular, we apply cross-modality adaptive meta-learning strategies to enable the model to identify and adapt to domain-invariant factors across different scenarios, significantly improving its generalization across diverse data environments. Conventional methods that rely solely on image enhancement techniques, such as rotation, often fail to ensure effective feature learning due to the potential for overfitting to specific enhancements, thereby reducing the model's ability to recognize unenhanced images. To address this, we introduce a dynamic error sample memory unit to store and reuse challenging hard negative samples, leveraging the fast adaptation and transfer learning capabilities of meta-learning. This approach enhances the model's ability to discern valid combinations of image and text features, leading to more accurate decisions in similar future scenarios. For optimization, we further introduce an adaptive dual-speed update strategy to balance fast adaptability and precise tuning. Fast updates allow the model to rapidly adapt to the basic features of new tasks, while slow updates focus on detailed parameter adjustments, ensuring stability and performance during long-term training. This enables the model to generalize effectively even with limited or incomplete text descriptions, making it well-suited for real-world applications.
In summary, our main contributions are as follows:
\begin{itemize}[label={},leftmargin=*,align=left]
    \item[$\bullet$] We introduce a Cross-modality Adaptive Meta-Learning (CAMeL) to facilitate domain-agnostic pretraining for text-based person retrieval, 
    which mitigates the impact of inherent domain biases in the generated data. 
    \item[$\bullet$] For multi-task optimization, we further propose a dynamic error sample memory unit and an adaptive dual-speed update strategy to balance the memorization of historical tasks and fast adaptation to new tasks. 
    \item[$\bullet$] Extensive experiments show that our 
    domain-agnostic pretraining framework via CAMeL has achieved a competitive recall rate on real-world benchmarks, \ie, CUHK-PEDES, ICFG-PEDES, and RSTPReid, surpassing existing methods. 
    Even for the ill-posed text query, \ie, missing several words, the proposed method still yields robust retrieval performance. 
\end{itemize}

\begin{figure*}[tp]
    \centering
    \includegraphics[width=1\linewidth]{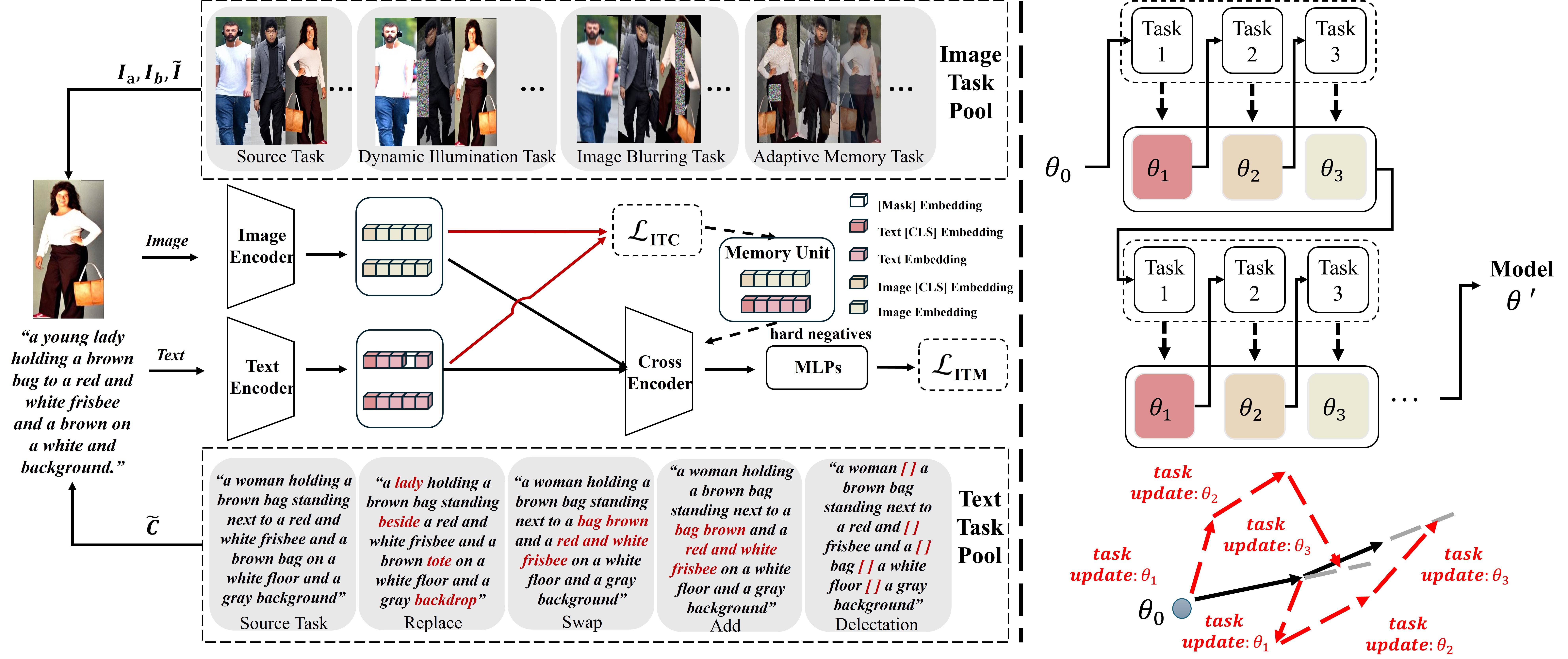}
    \caption{Overview of the proposed domain-agnostic pretraining on the synthetic dataset, \ie, MALS. 
    \textbf{(1)} We initially design stylized image tasks involving dynamic illumination, image blurring and adaptive memory, while we apply text augmentation to simulate the real-world natural language inputs. Then augmented image-text pairs are fed into the encoders, and calculate the image-text contrastive loss (ITC) and image-text matching loss (ITM).
    \textbf{(2)} Subsequently, guided by the meta-learning strategy, model parameters are optimized through gradient updates directed by the loss function, adapting to diverse task requirements. The red dashed line represents the task-specific updates, reflecting the model's rapid optimization in specific tasks (lines 6-10 in Alg.\ref{algorithm:1}). The gray dashed line represents the fast update, which helps the model quickly adapt to new tasks by adjusting global parameters (line 14 in Alg.\ref{algorithm:1}). The black line represents the slow update, ensuring gradual convergence through global optimization (line 17 in Alg.\ref{algorithm:1}).
    }
    \vspace{-.1in}
    \label{fig2}
\end{figure*}

The paper is organized as follows. Section~\ref{Sec.2} introduces some related works. Section~\ref{Sec.3} describes the domain-agnostic pretraining in detail. Section~\ref{Sec.4} discusses the experiment results conducted on some commonly used datasets. Section~\ref{Sec.5} concludes the paper and offers suggestions for future work. 

\section{Related Work}
\label{Sec.2}
\subsection{Text-to-Image Person Retrieval}  
\noindent Using natural language descriptions for person retrieval is more practical than relying solely on image or attribute queries. Li~\etal first propose text-to-image person retrieval and have created the large-scale descriptive dataset CUHK-PEDES~\cite{li2017person1}. With technological advancements and the diversification of application scenarios, researchers face challenges in accurately understanding and matching the complex relationships between textual descriptions and images. To advance retrieval technology, more complex datasets, such as ICFG-PEDES~\cite{ding2021semantically12} and RSTPReid~\cite{zhu2021dssl13}, have been introduced. The evolution from initial methods focusing on basic feature extraction (\eg, the GNA-RNN model for cross-modality data management~\cite{li2017person1}) to the current adoption of cross-modal attention mechanisms~\cite{shao2022learning32, bai2023rasa33} and deep learning frameworks~\cite{fujii2023bilma34} marks a significant transition. The introduction of attention mechanisms, particularly those utilizing human pose information to locate discriminative regions~\cite{wang2021dynamic21}, facilitates multi-granularity feature alignment between images and texts, while Wang~\etal~\cite{wang2023progressive} focus on the efficiency for the retrieval acceleration. Multi-granularity image-text alignment models by Niu~\etal~\cite{niu2020improving23} and the adversarial matching approach by Nikolaos~\etal~\cite{sarafianos2019adversarial24} further demonstrate the synergistic effects between data at various levels of detail. The VGSG method proposed by Ding~\etal~\cite{he2023vgsg25} achieves part-level feature alignment through semantic grouping without the need for additional pose alignment tools.
Recent studies further refine these approaches. Ergasti~\etal~\cite{ergasti2024mars} highlight the role of visual attributes in text-based person search, while Tan~\etal~\cite{tan2024occluded} propose a saliency-guided patch transfer method to address occlusion in person re-identification. Zhang~\etal~\cite{zhang2024adaptive} enhance cross-modal generalization via middle modality alignment for visible-infrared ReID, and Tan~\etal~\cite{tan2024rle} unify data augmentation strategies for cross-spectral ReID.
However, the above-mentioned methods do not explicitly deal with the domain gap between the pretraining and the target dataset. In this work, we do not pursue the final performance but focus more on the scalability of the pretrained model for different scenarios.

\subsection{Domain-agnostic Pretraining.} 
\noindent In recent years, with the advancement of machine learning, domain-agnosticism has become a research focus, particularly in reducing a model's dependency on specific domain knowledge and enhancing its generalization capabilities~\cite{verma2021towards66, tamkin2021dabs67, lee2020mix68}. This direction emphasizes developing models capable of learning data-related distortions during the pretraining phase. By leveraging a range of data sources, these models can acquire general features and patterns in pretraining, enabling rapid adaptation to unseen domains or tasks~\cite{mishra2022task2sim70, zhu2024mario}. Some studies align source domain features with target domain semantics to regularize cross-domain representation learning~\cite{huo2022domain71}, while the domain-agnostic prompting approach proposed by~\cite{du2024domain72} leverages domain-invariant semantics by aligning visual and textual embeddings.
Recent studies have shown that large multilingual models enhance zero-shot multimodal learning across languages~\cite{hu2023large}, while prompt learning research has explored the effect of uninformative class names on generalization~\cite{lv2024rethinking}.
In this study, we explore the use of domain-agnostic pretraining for text-to-image person retrieval and focus on multiple cross-modality tasks to learn domain-invariant features.

\subsection{Meta-learning}  
\noindent Meta-learning is widely used to improve generalization in cross-modal learning, helping models quickly adapt to new tasks with limited data. Its key strength lies in efficiently learning new tasks using methods like MAML and Reptile, enabling rapid adaptation with minimal data~\cite{finn2017model35, nichol2018reptile31, 10.1145/3589334.364536990}. For instance, Adaptive Uncertainty Learning (AUL) introduces an uncertainty-aware matching mechanism that leverages meta-learning to optimize cross-modal pairs, enhancing generalization on complex datasets~\cite{li2024adaptive83}. Meta-learning is effective in data-scarce scenarios, enabling rapid adaptation and improved performance~\cite{ma2022multimodality84, 10.1145/3543507.3583548, 10.1145/3485447.3512013}. Additionally, Meta-transfer Learning (MTL) addresses domain adaptation by learning meta-feature transformers that enhance adaptability to new domains, especially in unsupervised settings~\cite{sun2019meta85}. Meta-learning improves task performance on modality tasks, particularly in handling missing modalities~\cite{tran20233fm86}. Moreover, it has been applied to optimize multi-modal alignment, such as image-text alignment, by dynamically adjusting sample weights to prioritize high-quality samples, improving cross-modal alignment~\cite{wang2022look57, vettoruzzo2024advances88}. Different from previous works, we introduce meta-learning for domain-agnostic pretraining to minimize the domain gap while facilitating domain-invariant feature learning.

\section{Domain-agnostic Pretraining}
\label{Sec.3}
\noindent\textbf{Overview.} The objective of the domain-agnostic pretraining framework is to mitigate the adverse effects of biases between generated and real data on model training, thus enhancing model performance in text-based person retrieval tasks. 
In this section, we detail the proposed domain-agnostic pretraining framework for text-based person retrieval, which is depicted in Fig.~\ref{fig2}. 
Firstly, we utilize the generated dataset MALS~\cite{yang2023towards11} as our pretraining dataset, designing various tasks such as the Dynamic Illumination Task, Image Blurring Task and Adaptive Memory Task to simulate real-world data complexity. By employing meta-learning strategies, the model is enabled to identify and adapt to key variables across different cross-modality tasks, thereby enhancing its generalization capabilities across diverse data environments. Further, the cross-modal meta-learning strategy uses these varied cross-modality tasks to train the model, allowing fast adaptation to new cross-modality tasks in a short time. In our context, this means the model can learn from various types of data, discerning which information is useful and which could potentially lead to misguidedness. In this way, the model not only learns solutions for specific tasks but also acquires learning skills that can be transferred across different cross-modality tasks. Lastly, to further improve the capability of the model for in-depth exploration of individual tasks, we have introduced an adaptive dual-speed strategy. This allows the model to rapidly assimilate new knowledge while also deeply analyzing and optimizing long-term learning content. Fast updates provide an immediate response to new tasks, while slow updates ensure that more stable and accurate knowledge is refined from these responses, optimizing the long-term performance of the model. The algorithm is summarized in Alg.\ref{algorithm:1}.

\begin{algorithm}[t]
\caption{Cross-modality meta-learning with adaptive dual-speed updates}
\label{algorithm:1}
\begin{algorithmic}[1]

\Require Initial model parameters $\theta$, two meta-learning rate $\epsilon_{slow}$, $\epsilon_{fast}$, update cycle $k$, number of tasks $N$
\Ensure Trained model parameters $\theta'$
\State Initialize fast parameters $\theta_0 = \theta$
\State Initialize slow parameters $\theta' = \theta$
\For{each meta-epoch}
    \For{each cross-modality task $T_i$ in \{1, 2, ... , $N$\}}
        \State Initialize $\theta_i = \theta_{i-1}$
        \For{each training iterations within $T_i$}
            \State Sample augmented batch from $T_i$
            \State Compute loss $\mathcal{L}_{T_i}(\theta_i)$
            \PinkComment{Apply the sequential task-specific updates}
            \State Update task parameters $\theta_i$
        \EndFor
    \EndFor
    \PinkComment{Apply fast update to model parameters}
    \State Aggregate updates: $\theta_0 \gets \theta_0 + \epsilon_{fast}\frac{1}{N} \sum_{i=1}^{N} (\theta_i - \theta_0)$
    \If{meta-epoch \% $k == 0$}
        \PinkComment{Apply slow update to model parameters}
        \State Apply slow update: $\theta' \gets \theta' + \epsilon_{slow} (\theta_0 - \theta')$
        \State Set $\theta_0 = \theta' $
    \EndIf
\EndFor
\State \Return $\theta'$

\end{algorithmic}
\end{algorithm}



    
    


\subsection{Stylization Tasks}\label{sub3.1}
\noindent To enhance generalization capabilities across various application scenarios, a series of tasks have been designed that simulate the complexity and diversity of real-world data. Given the specifics of the dataset generation process, we emphasized image style variations such as lighting, contrast, and occlusion to train the model for diverse visual conditions. In text processing, a range of text transformation techniques have been utilized to simulate different image descriptions. These techniques include synonym replacement, random insertion, random deletion, and random swapping, creating an enriched set of text descriptions. By altering the original text probabilistically to mimic the annotation styles of different experts, the ability to comprehend and match diverse styles and expressions of text descriptions is improved.

\noindent\textbf{Dynamic Illumination Task}. Given images from the pretraining dataset MALS, represented as \(I\), we enhance the image dataset \(I_a\) by dynamically adjusting the illumination levels. Concurrently, with a certain probability, operations such as random rotation and cropping are applied to alter the visual representation of the images, simulating the appearance of images under different viewing angles and environmental illumination conditions.

\noindent\textbf{Image Blurring Task.} 
During the pretraining phase, we introduced a blurring task by applying Gaussian blur with varying intensities to the images in the set \(I\), generating a blurred image set \(I_b\). This process simulates common image quality issues, such as focus drift and motion blur. The goal of this task is to improve the ability of model to recognize images with partially missing or degraded visual information.
\begin{equation}
    I_b(x, y) = I(x, y) \odot G(x, y; \sigma),
    \label{equation1}
\end{equation}
where \(\odot\) is a convolution operation, \(G(x, y; \sigma)\) is a Gaussian fuzzy kernel as $ G(x, y; \sigma) = \frac{1}{2\pi \sigma^2} e^{-\frac{x^2 + y^2}{2\sigma^2}}$. \(\sigma \) is the standard deviation of the Gaussian kernel, which is used to control the intensity of the ambiguity. This operation not only challenges the ability of model to recognize images with partial loss of visual information but also simulates common image quality issues caused by inaccurate camera focus or motion blur.


\noindent\textbf{Adaptive Memory Task.} In text-based task retrieval, data augmentation methods~\cite{Wei_Zou_201927, Cubuk_Zoph_Shlens_Le_202028} such as rotation, flipping, or color adjustment provide some performance improvements by increasing data diversity. However, these static and predefined methods struggle to adapt to subtle variations in real-world scenarios. To address this issue, we introduce an Adaptive Memory Task that utilizes a dynamic erroneous sample memory unit. The dynamic hard negative sample memory unit takes images \(I_a\) and \(I_b\) produced by the Dynamic Illumination Task and Image Blurring Task as image inputs, \(C_a\) and \(C_b\) as textual caption inputs. It employs Mixup~\cite{han2022g30}  and memory bank~\cite{zhong2024memorybank41} mechanisms to simulate data variability. Samples from different tasks are dynamically integrated and updated to generate erroneous sample pairs and output real-time adjusted hard negatives. This enables the model to continually learn from and adapt to new and unknown sample features.

Specifically, by linearly interpolating between original input vectors \(I_a, I_b\) and their corresponding textual captions \(C_a, C_b\), new vectors \(\tilde{I}\) and new captions \(\tilde{C}\) are created, merging features from two different data samples to generate hard negative sample pairs. The mixing ratio in the Mixup process is controlled by the parameter \(\lambda\), which follows a Beta distribution, \ie, \(\lambda \sim \text{Beta}(\delta, \delta)\), ensuring a balanced contribution of two sets of tasks during the mixing process. After undergoing two types of augmentations tasks, we obtain two augmented images and captions, \(I_{\text{a}}\), \(I_{\text{b}}\) and \(C_{\text{a}}\), \(C_{\text{b}}\). The formulas for the mix can be represented as:

\begin{equation}
    \tilde{I} = \lambda \cdot I_{\text{a}} + (1 - \lambda) \cdot I_{\text{b}},
    \tilde{C} = \lambda \cdot C_{\text{a}} + (1 - \lambda) \cdot C_{\text{b}}.
    \label{equation3and4}
\end{equation}
Therefore, we could define three typical tasks $T_i$. $T_1$ consists of \(I_{\text{a}}\) and \(\tilde{C}\), while $T_2$ is composed of \(I_{\text{b}}\) and \(\tilde{C}\). $T_3$ contains both \(\tilde{I}\) and \(\tilde{C}\).

\subsection{Dynamic Error Sample Memory Unit}
\noindent The memory unit mechanism plays a crucial role in storing and dynamically updating historical information from previous tasks, enabling the model to randomly sample erroneous samples for replaying when learning new tasks. However, this method requires substantial working memory, which is usually infeasible, particularly in text-based pedestrian retrieval tasks where precise matching of images and corresponding text descriptions is necessary. 
To address this challenge, we focus specifically on identifying and replaying those error samples that pose significant challenges to model performance improvement, known as ``hard negatives."
As shown in Fig.~\ref{fig2}, when introducing a new batch of hard negatives (\eg, \(\tilde{I}\), \(\tilde{C}\)), if the memory unit is not full, we directly add the embedding of these new samples. If it is full, we first remove some of the old hard negative samples from the unit and then add the new samples in a queue format. This update strategy not only ensures that the data in the memory unit is continuously updated, maintaining its immediacy and relevance to the training process, increasing the difficulty of the image-text matching (ITM)~\cite{wang2022point73} task, but also effectively reduces the required storage space by discarding old samples that no longer contribute to learning.
Moreover, this approach enhances the ability of model to recognize fine-grained features and classify intra-class variations, effectively simulating the variety of changes encountered in natural environments. Albiet simple, this strategy not only enriches the training batches but also improves the generalization ability of model across different scenarios, significantly enhancing its robustness and  
consistency in historical tasks. 

\subsection{Cross-modality Meta-learning} 
\noindent In text-based person retrieval tasks, the model must simultaneously handle data from two distinct modalities: images and text. This requires the model not only to capture information from each individual modality but also to understand and utilize the correlations between them. Therefore, we integrate information from different modalities and employ a cross-modality adaptive meta-learning approach, enabling the model to quickly adapt to new tasks. Meta-learning~\cite{finn2017model35,nichol2018first76,ravi2016optimization77,wu2019ace78,nichol2018reptile31} has been proven to excel in rapid task adaptation, allowing the model to adjust swiftly to new challenges through simple SGD~\cite{robbins1951stochastic82} updates. Considering that the model will undergo fine-tuning on new tasks using a gradient-based approach, our learning objective is to enable rapid adaptation to new tasks based on the extraction of internal features, while avoiding over-fitting. This approach ensures rapid progress on new tasks while maintaining generalization capabilities. Specifically, 
multiple meta-learning epochs are conducted, with each epoch consisting of several tasks. 
For each cross-modality task $T_i$, we initialize the task-specific parameters to the last parameters $\theta_i = \theta_{i-1}$. 
When $i=0$, we set $\theta_0$ as the initial model parameter $\theta$.
Then, $\theta_i$ are updated through iterations of gradient descent of the cross-modality task $T_i$ as:
\begin{equation}
\theta_i = \theta_{i} - \eta \nabla \mathcal{L}_{T_i}(\theta_{i}), 
\label{equation6}
\end{equation}
where \( \nabla \mathcal{L}_{T_i}(\theta_i) \) is the gradient of the loss function \( \mathcal{L}_{T_i} \) with respect to the parameters \( \theta_i \) for task \( T_i \), and $\eta$ is the learning rate used for each cross-modality task.
Upon completion of all tasks, the model aggregates these updates to obtain fast parameters $\theta_0$ as follows:
\begin{equation}
   \theta_0 = \theta_0 + \epsilon_{fast}\frac{1}{N} \sum_{i=1}^{N}(\theta_i - \theta_0 ), 
   \label{equation5}
\end{equation}
where \( \epsilon_{fast} \) is the fast meta-learning rate to aggregate task updates, and $N$ is the number of tasks.

\subsection{Adaptive Dual-Speed Update} 
\noindent To effectively implement cross-modality adaptive meta-learning and optimize performance in neural networks, substantial effort is often required to adjust hyperparameters. Particularly in the detailed exploration of independent tasks, to more precisely master the nuances of each task, we have adopted an adaptive dual-speed update strategy. During the fast update phase, the model iterates swiftly by exploring potential search directions. Subsequently, based on the data provided by these fast iterations, the slow update phase integrates this information to optimize and confirm the final direction of model, resulting in a more robust and efficient optimization process.
Specifically, at the beginning of model training, we duplicate the model parameters, creating two sets: one for fast updates (denoted as \( \theta_0 \)), and another for slow updates (denoted as \( \theta' \)), with the initial parameters set as \( \theta \). For individual tasks in meta-learning, we use \( \theta_i \) for regular training optimization. However, after every \( k \) iterations of task training, the slow parameters \( \theta' \) is updated using linear interpolation between \( \theta_0 \) and \( \theta' \). The rule for fast updates \( \theta_0 \) is presented in Equation~\ref{equation5}.
Following this, the slow updates are made based on the state after \( k \) task iterations. The slow update rule is as follows:
\begin{equation}
   \theta' =  \theta' \ + \epsilon_{slow} (\theta_0 - \theta'),
   \label{equation8}
\end{equation}
where \( \epsilon_{slow} \) is the slow meta-learning rate to control the impact of fast updates on the parameters of the model, confirm the final update direction of model.

\section{Experiment}
\label{Sec.4}
\noindent In this section, we provide a detailed description of the experimental part, which is divided into three main subsections: experimental setup, comparison with competitive methods, and discussion.
We begin by introducing three large-scale image-text retrieval datasets, CUHK-PEDES, ICFG-PEDES and RSTPReid, followed by the evaluation metrics, and then detail the implementation specifics. The comparison with competitive methods and thoughts on the experimental study will be discussed in Sections~\ref{sub4.2} and~\ref{sub4.3}, respectively.

\subsection{Experimental Setup}

\noindent\textbf{Datasets.} 
We evaluated our method on three challenging text-to-image person retrieval datasets. \textbf{CUHK-PEDES}~\cite{li2017person1} contains 40,206 images and 80,412 descriptions corresponding to 13,003 identities, with splits of 11,003 for training, 1,000 for validation, and 1,000 for testing. \textbf{RSTPReid}~\cite{zhu2021dssl13} is created by compiling MSMT-17~\cite{wei2018person42} data, includes 20,505 images and 41,010 descriptions for 4,101 individuals, divided into 3,701 identities for training and 400 for testing. \textbf{ICFG-PEDES}~\cite{ding2021semantically12} is also derived from MSMT-17 and consists of 54,522 images and descriptions for 4,102 identities, with 3,102 identities for training and 1,000 for testing.

\begin{table}[t]
\caption{\textbf{Performance Comparison on CUHK-PEDES.} 
Baseline (Pretrained): The same model architecture as CAMeL, pre-trained on the same dataset, but without the incorporation of ST, CMML, and ADSU. Baseline (Finetuned): The Baseline model (Pretrained) further fine-tuned on the target dataset. CAMeL (Pretrained): The pre-trained model applied directly to the target dataset without fine-tuning. CAMeL (Finetuned): The CAMeL (Pretrained) model further fine-tuned on the target dataset. $\dagger$: only reports the number of trainable parameter.}
\centering
\small
\sisetup{table-format=2.2} 
\resizebox{\linewidth}{!}{%
\begin{tabular}{l|ccccc}
\toprule
Method & {\#Parameter}& {R1} & {R5} & {R10} & mAP \\
\midrule
Dual Path~\cite{zheng2020dual8} &{-}& 44.40 & 66.26 & 75.07 & {-} \\
CMPM+CMPC~\cite{zhang2018deep7} &{-}& 49.37 & {-} & 79.21 & {-} \\
MIA~\cite{niu2020improving23} &{-}& 53.10 & 75.00 & 82.90 & {-} \\
A-GANet~\cite{liu2019deep48} &{-}& 53.14 & 74.03 & 81.95 & {-} \\
ViTAA~\cite{wang2020vitaa49} &{177M}& 55.97 & 75.84 & 83.52 & 51.60 \\
IMG-Net~\cite{wang2020img50} &{-}& 56.48 & 76.89 & 85.01 & {-} \\
CMAAM~\cite{aggarwal2020text51} &{-}& 56.68 & 77.18 & 84.86 & {-} \\
HGAN~\cite{zheng2020hierarchical52} &{-}& 59.00 & 79.49 & 86.62 & {-} \\
NAFS~\cite{gao2021contextual53} &{189M}& 59.94 & 79.86 & 86.70 & 54.07 \\
DSSL~\cite{zhu2021dssl13} &{-}& 59.98 & 80.41 & 87.56 & {-} \\
MGEL~\cite{wang2021text54} &{-}& 60.27 & 80.01 & 86.74 & {-} \\
SSAN~\cite{ding2021semantically12} &{-}& 61.37 & 80.15 & 86.73 & {-} \\
NAFS~\cite{gao2021contextual53} &{189M}& 61.50 & 81.19 & 87.51 & {-} \\
TBPS~\cite{han2021text55} &{43M}& 61.65 & 80.98 & 86.78 & {-} \\
TIPCB~\cite{chen2022tipcb56} &{185M}& 63.63 & 82.82 & 89.01 & {-} \\
LBUL~\cite{wang2022look57} &{-}& 64.04 & 82.66 & 87.22 & {-} \\
CAIBC~\cite{wang2022caibc58} &{-}& 64.43 & 82.87 & 88.37 & {-} \\
AXM-Net~\cite{farooq2022axm59} &{-}& 64.44 & 80.52 & 86.77 & 58.73 \\
SRCF~\cite{suo2022simple60} &{-}& 64.88 & 83.02 & 88.56 & {-} \\
LGUR~\cite{shao2022learning32} &{-}& 65.25 & 83.12 & 89.00 & {-} \\
CFine~\cite{yan2023clip61} &{-}& 69.57 & 85.93 & 91.15 & {-} \\
PLIP-RN50~\cite{zuo2023plip62} &{-}& 69.23 & 85.84 & 91.16 & {-} \\
IRRA~\cite{jiang2023cross6} &{194M}& 73.38 & 89.93 & 93.71 & 66.13 \\
TBPS-CLIP~\cite{cao2024empirical63} &{149M}& 73.54 & 88.19 & 92.35 & 65.38 \\
RDE~\cite{qin2023noisy} &{153M}& 75.94 & 90.63 & 94.12 & 67.56 \\
RaSa~\cite{bai2023rasa33} &{210M}& 76.51 & 90.29 & 94.25 & \textbf{69.38} \\
APTM~\cite{yang2023towards11} &{214M}& 76.53 & 90.04 & 94.15 & 66.91 \\
WoRA~\cite{sun2024data} &{127M$^\dagger$}& 76.38 & 89.72 & 93.49 & 67.22 \\
\midrule
Baseline (Pretrained) &{145M}& 15.97 & 30.90 & 40.22 & 14.77 \\
Baseline (Finetuned) &{145M}& 74.58 & 88.97 & 93.63 & 65.56 \\
CAMeL (Pretrained) &{145M}& 25.26 & 44.04 & 52.62 & 22.52 \\
CAMeL (Finetuned)  &{145M}& \textbf{77.24} & \textbf{91.80} & \textbf{95.16} & 68.32 \\
\bottomrule
\end{tabular}
} 
\label{tab1}
\end{table}

\noindent\textbf{Evaluation Metrics.} To comprehensively evaluate the effectiveness of our proposed model, following previous practices, we adopted Recall@K (where K is 1, 5, 10) as our primary evaluation metric. This metric reflects the ability of the model to successfully identify the target image of the people among the top k most relevant image candidates upon receiving a specific text query. Furthermore, as a supplement to assess the overall retrieval performance of a model, we also introduced the mAP as an evaluation metric. Within this evaluation framework, higher values of Recall@K and mAP indicate superior model performance.

\noindent\textbf{Implementation Details.} 
Our model comprises three encoders: an \textit{image encoder} initialized with $12$ layers of SG-Former~\cite{ren2023sg38}, a \textit{text encoder} initialized with the first $6$ layers of BERT~\cite{devlin2018bert18}, and a \textit{cross encoder} initialized with the last $6$ layers of BERT.
For image augmentation, we employed adaptive techniques such as RandAugment~\cite{Cubuk_Zoph_Shlens_Le_202028}, using random horizontal flipping, random erasing, and random cropping, with all images resized to 224×224 pixels. For text augmentation, the probabilities for synonym replacement, random insertion, swapping, and deletion were set at 10\% for each token. The maximum text sequence length was set to 56, with an embedding dimension of 256.
The hyperparameter \( \lambda = \text{np.random.beta}(\delta, \delta) \) with \( \delta = 1 \). The Baseline (Pretrained) model is trained solely on the synthetic dataset MALS without any fine-tuning on the target dataset, whereas the Baseline (Fine-tuned) model undergoes additional training on the target dataset to better adapt to its distribution. Notably, the augmentation strategies are consistently applied across all experiments to ensure comparability.

Model training proceeds in two phases: pretraining and fine-tuning. During pretraining, we train for 32 epochs on four NVIDIA A6000 GPUs, with a batch size of 80. We use AdamW~\cite{loshchilov2017decoupled43} as the optimizer, with an initial learning rate of 1e-4 using linear decay and a weight decay of 0.01. 
Following pretraining, the model undergoes fine-tuning on downstream datasets for 30 epochs, with each session lasting around five hours. During the first 10 epochs, the focus is on stylized tasks to enhance data understanding, while the remaining 20 epochs transition into a stable optimization phase, where the learning rate is fixed at 2e-4 for task-specific fine-tuning.
A slow update is applied after completing every six cross-modality tasks. 
Additionally, every 3 epochs, we capture a snapshot of the model's weight parameters and incorporate it into the stochastic weight averaging process~\cite{izmailov2018averaging44}, balancing generalization and adaptability across cross-modality tasks. We clarify that the model fine-tuning process takes approximately 5 hours. Moreover, the inference time per text query of ours is 5.6ms, validating efficiency comparable to the APTM method.

\subsection{Comparison with Competitive Methods}\label{sub4.2}
\noindent We compare our method with state-of-the-art text-based person retrieval models on CUHK-PEDES, ICFG-PEDES, and RSTPReid datasets, as detailed in Tables~\ref{tab1},~\ref{tab2}, and~\ref{tab3}. Our proposed model consistently outperforms existing methods across all three datasets, achieving a competitive recall rate while significantly reducing the number of computational parameters.

On CUHK-PEDES, our method achieves a Recall@1 of 77.24\%, Recall@5 of 91.80\%, and Recall@10 of 95.16\%, with an mAP of 68.32\%. In contrast, the second-best method, APTM, records a Recall@1 of 76.53\% and an mAP of 66.91\%. The consistent improvement, particularly a 0.71\% increase in Recall@1, highlights our model's ability to better align fine-grained textual descriptions with image representations. This improvement is attributed to the introduction of cross-modality adaptive meta-learning (CAMeL), which not only enhances generalization across diverse data environments, but also significantly reduces computational complexity through more efficient parameter utilization.
Notably, despite the significant reduction in parameters, our model achieves approximately a 5\% improvement in every performance metric compared to the TBPS-CLIP model with a similar number of parameters on the CUHK-PEDES dataset, and maintains the same level of improvement across other datasets. This validates that our model delivers excellent performance without increasing computational overhead.

For RSTPReid, which focuses on person re-identification across real-world surveillance scenarios with challenging occlusions and viewpoint variations, our method validates an average improvement of 1.25\% across key metrics. Importantly, our model’s ability to handle difficult conditions in occluded or partially visible scenes is a direct result of the dynamic error sample memory unit, which improves robustness in such scenarios.

On ICFG-PEDES, our method achieves a Recall@1 of 68.70\% and an mAP of 41.58\%, outperforming the previous best model in all metrics. The robust performance on this dataset further validates the capability of our adaptive dual-speed update (ADSU) strategy to balance fast adaptation with stable, long-term learning, ensuring our model's competitiveness in cross-modal person retrieval tasks.
These results validate the effectiveness of our method, not only setting new benchmarks but also demonstrating its robustness and adaptability across diverse real-world scenarios. The proposed approach consistently outperforms state-of-the-art methods, allowing it to generalize effectively even in challenging settings.

\begin{table}[t]
\caption{\textbf{Performance Comparison on RSTPReid.}}
\centering
\small
\sisetup{table-format=2.2} 
\resizebox{\linewidth}{!}{%
\begin{tabular}{
  l|
  c
  c
  c
  c
  c
}
\toprule
Method & {\#Parameter}& {R1} & {R5} & {R10} & mAP \\
\midrule
DSSL~\cite{zhu2021dssl13} &{-}& 32.43 & 55.08 & 63.19 & {-} \\
LBUL~\cite{wang2022look57} &{-}& 45.55 & 68.20 & 77.85 & {-} \\
IVT~\cite{shu2022see40} &{-}& 46.70 & 70.00 & 78.80 & {-} \\
CAIBC~\cite{wang2022caibc58} &{-}& 47.35 & 69.55 & 79.00 & {-} \\
CFine~\cite{yan2023clip61} &{-}& 50.55 & 72.50 & 81.60 & {-} \\
IRRA~\cite{jiang2023cross6} &{194M}& 60.20 & 81.30 & 88.20 & 47.17 \\
TBPS-CLIP~\cite{cao2024empirical63} &{149M}& 61.96 & 83.55 & 88.75 & 48.26 \\
RDE~\cite{qin2023noisy} &{153M}& 65.35 & 83.95 & 89.9 & 50.88 \\
RaSa~\cite{bai2023rasa33} &{210M}& 66.90 & 86.50 & 91.35 & 52.31 \\
APTM~\cite{yang2023towards11} &{214M}& 67.50 & 85.70 & 91.45 & 52.56 \\
WoRA~\cite{sun2024data} &{127M$^\dagger$}& 66.85 & 86.45 & 91.10 & 52.49 \\
\midrule
Baseline (Pretrained) &{145M}& 20.90 & 43.60 & 54.80 & 15.50 \\
Baseline (Finetuned) &{145M}& 67.15 & 86.80 & 91.95 & 52.93 \\
CAMeL (Pretrained) &{145M}& 26.60 & 50.00 & 60.40 & 20.51 \\
CAMeL (Finetuned) &{145M}& \textbf{68.50} & \textbf{87.40} & \textbf{92.70} & \textbf{53.61} \\
\bottomrule
\end{tabular}
} 
\label{tab2}
\end{table}

\begin{table}[t]
\caption{\textbf{Performance Comparison on ICFG-PEDES.}}
\centering
\small
\sisetup{table-format=2.2} 
\resizebox{\linewidth}{!}{%
\begin{tabular}{
  l|ccccc
}
\toprule
Method & {\#Parameter}& {R1} & {R5} & {R10} & mAP \\
\midrule
Dual Path~\cite{zheng2020dual8} &{-}& 38.99 & 59.44 & 68.41 & {-} \\
MIA~\cite{niu2020improving23} &{-}& 46.49 & 67.14 & 75.18 & {-} \\
ViTAA~\cite{wang2020vitaa49} &{-}& 50.98 & 68.79 & 75.78 & {-} \\
SSAN~\cite{ding2021semantically12} &{-}& 54.23 & 72.63 & 79.53 & {-} \\
IVT~\cite{shu2022see40} &{-}& 56.04 & 73.60 & 80.22 & {-} \\
LGUR~\cite{shao2022learning32} &{-}& 59.02 & 75.32 & 81.56 & {-} \\
CFine~\cite{yan2023clip61} &{-}& 60.83 & 76.55 & 82.42 & {-} \\
IRRA~\cite{jiang2023cross6} &{194M}& 63.46 & 80.25 & 85.82 & 38.06 \\
TBPS-CLIP~\cite{cao2024empirical63} &{149M}& 65.05 & 80.34 & 85.47 & 39.83 \\
RDE~\cite{qin2023noisy} &{153M}& 67.68 & 82.47 & 87.36 & 40.06 \\
RaSa~\cite{bai2023rasa33} &{210M}& 65.28 & 80.40 & 85.12 & 41.29 \\
APTM~\cite{yang2023towards11} &{214M}& 68.51 & 82.99 & 87.56 & 41.22 \\
WoRA~\cite{sun2024data} &{127M$^\dagger$}& 68.35 & 83.10 & 87.53 & \textbf{42.60} \\
\midrule
Baseline (Pretrained) &{145M}& 11.81 & 25.28 & 32.99 & 3.57 \\
Baseline (Finetuned) &{145M}& 66.40 & 81.49 & 86.73 & 39.55 \\
CAMeL (Pretrained) &{145M}& 17.50  & 33.03 & 41.04 & 6.06 \\
CAMeL (Finetuned) &{145M}& \textbf{68.70} & \textbf{83.11} & \textbf{88.32} & 41.58 \\
\bottomrule
\end{tabular}
} 
\label{tab3}
\end{table}

\begin{table*}[t]
\caption{\textbf{Ablation study on each component of CAMeL on three benchmark datasets.} To minimize the impact of experimental randomness on our research contributions, we employed a method of averaging the results from ten trials to present our results.
Baseline indicates pretraining and fine-tuning with the same dataset and model structure, but not employing the proposed CAMeL;
ST: stylization tasks; ADSU: adaptive dual-speed update; CMML: cross-modal meta-learning.}
\centering
\small
\resizebox{\textwidth}{15mm}{
\begin{tabular}{
    c| 
    l| 
    c 
    c 
    c| 
    S[table-format=2.2] 
    S[table-format=2.2] 
    S[table-format=2.2] 
    S[table-format=2.2]| 
    S[table-format=2.2] 
    S[table-format=2.2] 
    S[table-format=2.2] 
    S[table-format=2.2]| 
    S[table-format=2.2] 
    S[table-format=2.2] 
    S[table-format=2.2] 
    S[table-format=2.2] 
}
\toprule
\multirow{2}{*}{No.} & \multirow{2}{*}{Methods} & \multicolumn{3}{c|}{Components} & \multicolumn{4}{c|}{CUHK-PEDES} & \multicolumn{4}{c|}{RSTPReid} & \multicolumn{4}{c}{ICFG-PEDES} \\
\cline{3-17}
& &  {ST} & {ADSU} & {CMML} & {R1} & {R5} & {R10} & {mAP} &{R1} & {R5} & {R10} & {mAP} & {R1} & {R5} & {R10} & {mAP}\\
\midrule
1 & Baseline & & & & 74.58 & 88.97 & 93.63 & 65.56 & 67.15 & 85.80 & 91.25 & 52.93 & 66.40 & 81.49 & 86.73 & 39.55 \\
2 & +ST & \checkmark & & & 75.05 & 89.86 & 94.12 & 66.00 & 67.50 & 86.50 & 91.95 & 53.14 & 67.18 & 81.85 & 86.90& 39.87 \\
3 & +ADSU & \checkmark & \checkmark &  & 75.42 & 90.29 & 94.19 & 66.53 & 67.95 & 86.90 & 92.05 & 52.91 &  67.64 & 82.33 & 87.21 & 40.56 \\
4 & +CMML & \checkmark & & \checkmark & 76.33 & 90.58 & 94.61 & 67.50  & 68.30 & 87.00 & 92.40 & 53.43& 68.01 & 82.54 & 87.59 & 40.80 \\ 
\midrule
5 & CAMeL & \checkmark & \checkmark & \checkmark & \textbf{77.24} & \textbf{91.80}  & \textbf{95.16} & \textbf{68.32} & \textbf{68.50} & \textbf{87.40} & \textbf{92.70} & \textbf{53.61} &\textbf{68.70} & \textbf{83.11} & \textbf{88.32} & \textbf{41.58} \\
\bottomrule
\end{tabular}}
\label{tab4}
\end{table*}

\begin{figure}[t]
    \centering
    \includegraphics[width=1\linewidth]{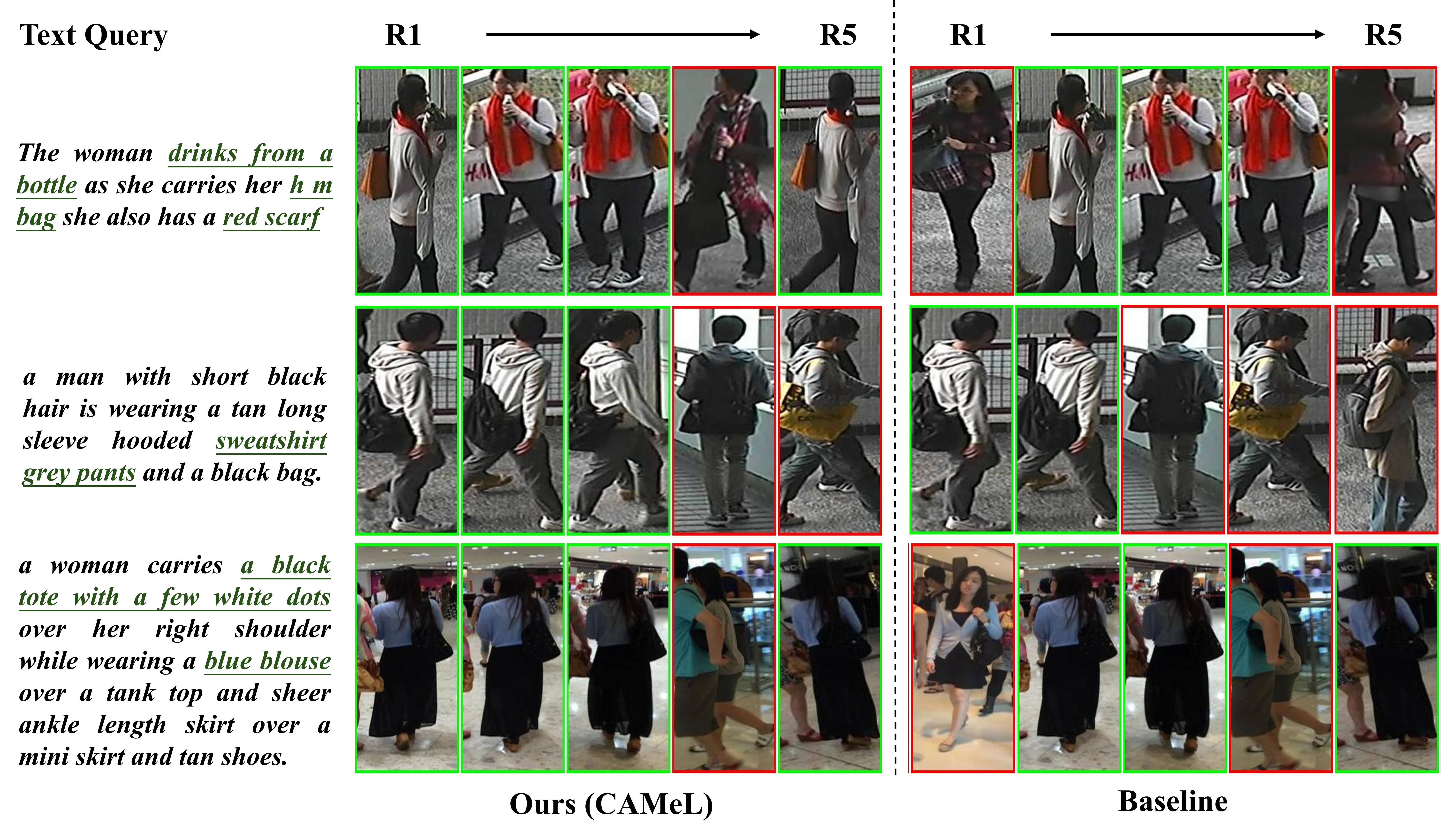}
    \caption{Qualitative comparison of text-to-image retrieval results between Ours (CAMeL) and the Baseline on the benchmark datasets, with results ordered by similarity from highest to lowest, left to right. Correct matches are highlighted with a green frame, while incorrect matches are marked in red. The green-highlighted text emphasizes the details accurately captured by our approach.}
    \vspace{-.1in}
    \label{fig5}
\end{figure}

\subsection{Ablation Studies and Further Discussion}\label{sub4.3}

\noindent\textbf{Effectiveness of Domain-agnostic Pretraining.}  
We design two sets of ablation studies: the first Baseline (Pretrained) indicates the same model structure with the same pretraining dataset and involves no fine-tuning and directly evaluation on the target tasks; the second acting as CAMeL (Pretrained), which also involves no fine-tuning and directly evaluation on the target tasks.
All experiments are carried out under the same dataset and task settings to ensure fairness and comparability of the results. 
As shown in Table~\ref{tab1}, models that undergo domain-agnostic pretraining significantly outperform Baseline on all evaluation metrics, underscoring the necessity of domain-agnostic pretraining. 
Models that do not undergo effective domain-agnostic learning not only perform poorly during the pretraining phase but also negatively impact the results of subsequent fine-tuning.
Fig.~\ref{fig5} presents a comparison of retrieval results between our method (CAMeL) and the baseline on the benchmark datasets.

\begin{figure}[t]
    \centering
    \includegraphics[width=0.95\linewidth]{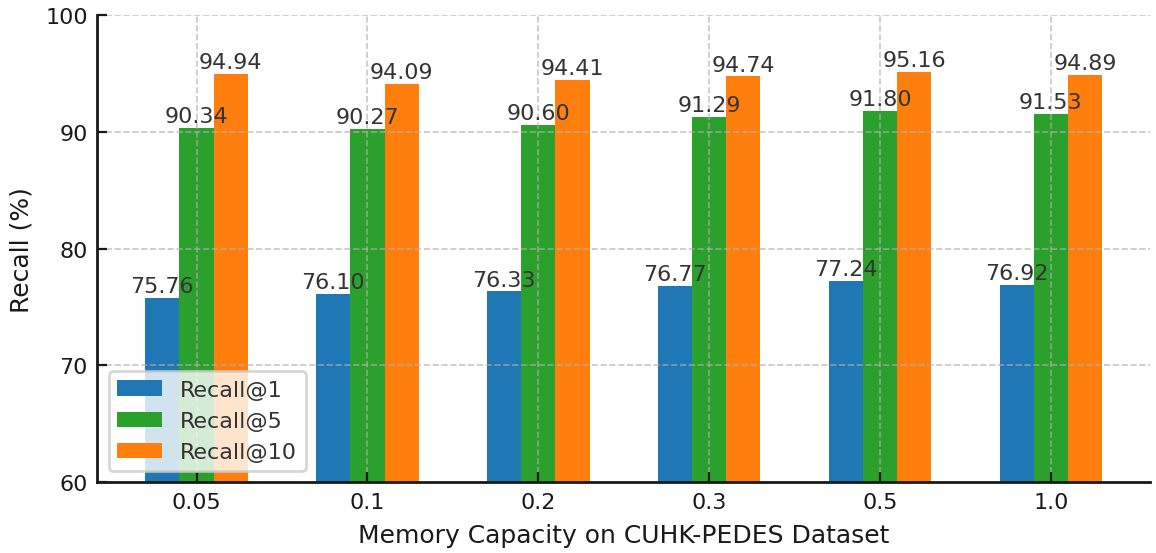}
    \caption{Ablation Study on the memory capacity in our CAMeL.
    We apply 5\%, 10\%, 20\%, 30\%, 50\% and 100\% data pairs to pre-train, and then report the fine-tuned performance on CUHK-PEDES dataset. 
    The percentage refers to the current capacity relative to the sample size extracted for dynamic illumination and blurring tasks.}
    \vspace{-.1in}
    \label{fig7}
\end{figure}

\noindent\textbf{Effectiveness of Stylization Tasks.}  
Considering the importance of pretraining, our experimental approach is based on the Baseline. We have conducted several quantitative experiments to validate the effectiveness of our stylized tasks (ST), as shown in Table~\ref{tab4}. By comparing the performance on the Baseline with and without ST, we find that simulating the complexity and diversity of real-world data through ST beneficially influences the learning of model parameters, leading to superior results across three benchmark datasets. Furthermore, to deepen our understanding of the impact of sample library size in adaptive memory tasks, we carry out a series of ablation experiments with sample library sizes ranging from 5\% to 100\% of the current sample batch. The results indicate that optimal model performance is achieved when the sample library size reaches 50\%, as shown in Fig.~\ref{fig7}. This is likely because a smaller sample library usually does not adequately cover the diversity of tasks, whereas a larger library introduces redundant information, increasing learning complexity and noise, thereby affecting the model's generalization ability. By optimizing the size of the sample library, we improve the model adaptability to complex scenarios while avoiding overfitting, ensuring the stability and efficiency of the model across various tasks.
Moreover, we further investigate the effect of task scheduling in the proposed meta-learning strategy. In our implementation, the model follows a fixed task order: it first performs the Dynamic Illumination Task, then the Image Blurring Task, and finally the Adaptive Memory Task. As shown in Tab.~\ref{tab:performance_comparison} results across three benchmark datasets show that the task order has negligible impact on final performance. This observation can be explained by the pronounced domain gap between real and synthetic data, as discussed in the Introduction. The three tasks are explicitly designed to simulate key domain characteristics such as lighting conditions, color consistency, and resolution. Their diversity and complementarity reduce sensitivity to task order, ensuring robust generalization across varied data distributions.

\begin{table}[t]
    \centering
    \caption{\textbf{Ablation study of tasks order on the three benchmark datasets.} Fixed(Now): Fixed order. Random: Random order.}
    \resizebox{\linewidth}{!}{%
    \begin{tabular}{c|c|c|c|c}
        \hline
        \textbf{CUHK-PEDES} & R1 & R5 & R10 & mAP \\ \hline
        Fixed{ (Now)} & 77.24 & \textbf{91.80} & 95.16 & 68.32 \\ 
        Random & \textbf{77.26} (+0.02) & 91.54 (-0.26) & \textbf{95.50} (+0.34) & \textbf{68.46} (+0.14) \\ \hline
        \textbf{ICFG-PEDES} & R1 & R5 & R10 & mAP \\ \hline
        Fixed{ (Now)} & \textbf{68.70} & 83.11 & \textbf{88.32} & \textbf{41.58} \\ 
        Random & 68.46 (-0.24) & \textbf{83.27} (+0.16) & 88.06 (-0.26) & 41.31 (-0.27) \\ \hline
         \textbf{RSTPReid} & R1 & R5 & R10 & mAP \\ \hline
        Fixed{ (Now)} & \textbf{68.50} & \textbf{87.40} & \textbf{92.70} & \textbf{53.61} \\ 
        Random & 68.40 (-0.1) & 86.95 (-0.45) & 92.55 (-0.15) & 53.27 (-0.34) \\ \hline
    \end{tabular}}
    \vspace{-.1in}
    \label{tab:performance_comparison}
\end{table}

\noindent\textbf{Effectiveness of Adaptive Dual-Speed Update.} 
To more precisely master the details of individual tasks, we introduce Adaptive Dual-Speed Update (ADSU). As shown in Table~\ref{tab4}, we establish control groups by comparing the sole use of ST with the addition of ADSU across three benchmark datasets, all of which demonstrate superior outcomes. This confirms the effectiveness of the dual-speed update strategy in enhancing the adaptability and stability of the model. Additionally, we explore the impact of various combinations of fast and slow update steps on model performance. We find that setting the update frequency to perform a slow update every six iterations for the three stylized tasks, coupled with a smoothing factor of 0.5, achieves optimal performance. As shown in Tab.~\ref{tab:performance_k}, ablation studies on the RSTPReid dataset further validate the effectiveness of this configuration. When \(k = 6\), the model achieves optimal results in Recall@1, Recall@5, Recall@10 and mAP, reaching 53.61\% mAP and 68.50\% Recall@1. In contrast, significantly smaller or larger \(k\) values disrupt the balance between fast and slow updates, thereby degrading overall performance. This allows the model to rapidly respond to the learning needs of new tasks while maintaining sufficient information accumulation, ensuring that slow updates can proceed after a more comprehensive evaluation of the accumulated learning outcomes.

\begin{table}[t]
    \centering
    \caption{\textbf{Ablation study of hyperparameter k on the RSTPReid datasets.}}
    \resizebox{\linewidth}{!}{%
    \begin{tabular}{c|c|c|c|c}
        \hline
        \textbf{RSTPReid} & R1 & R5 & R10 & mAP \\ \hline
        k=3  & 68.20 (-0.30) & 86.95 (-0.45) & 92.00 (-0.70) & 52.76 (-0.85) \\ 
        k=6 (Now) & \textbf{68.50} & \textbf{87.40} & \textbf{92.70} & \textbf{53.61} \\ 
        k=15  & 68.50 (-0.00) & 86.35 (-1.05) & 91.50 (-1.20) & 52.54 (-1.07) \\ 
        k=30  & 67.75 (-0.75) & 87.00 (-0.40) & 92.00 (-0.70) & 53.16 (-0.45) \\ 
        \hline
    \end{tabular}}
    \vspace{-.1in}
    \label{tab:performance_k}
\end{table}

To further validate the generalization capability of ADSU, we conduct experiments on a cross-modal geo-localization task under diverse weather conditions. This task involves two settings: drone-to-satellite and satellite-to-drone. Specifically, we apply ADSU to a dual-branch neural network and evaluate it on three datasets: University-1652~\cite{Zheng_Wei_Yang_202001}, SUES-200~\cite{zhu2023sues18}, and CVUSA~\cite{Zhai_Bessinger_Workman_Jacobs_201717}. In the drone-to-satellite task, the ADSU-enhanced method achieves Recall@1 scores of 66.88\%, 55.07\%, and 76.26\% on these datasets, respectively, outperforming state-of-the-art methods, which score 65.15\%, 52.02\%, and 75.00\%. Notably, under extreme weather conditions such as dark+rain, ADSU demonstrates a 6\% improvement. Similar improvements are observed in terms of Average Precision (AP). Consistent performance gains are also observed in the satellite-to-drone task, further underscoring the robustness of ADSU in viewpoint alignment tasks. These results highlight the significant performance improvements that ADSU achieves, particularly under challenging weather conditions, and validate its effectiveness and robustness in cross-modal localization tasks.

\begin{figure}[t]
    \centering
    \includegraphics[width=1\linewidth]{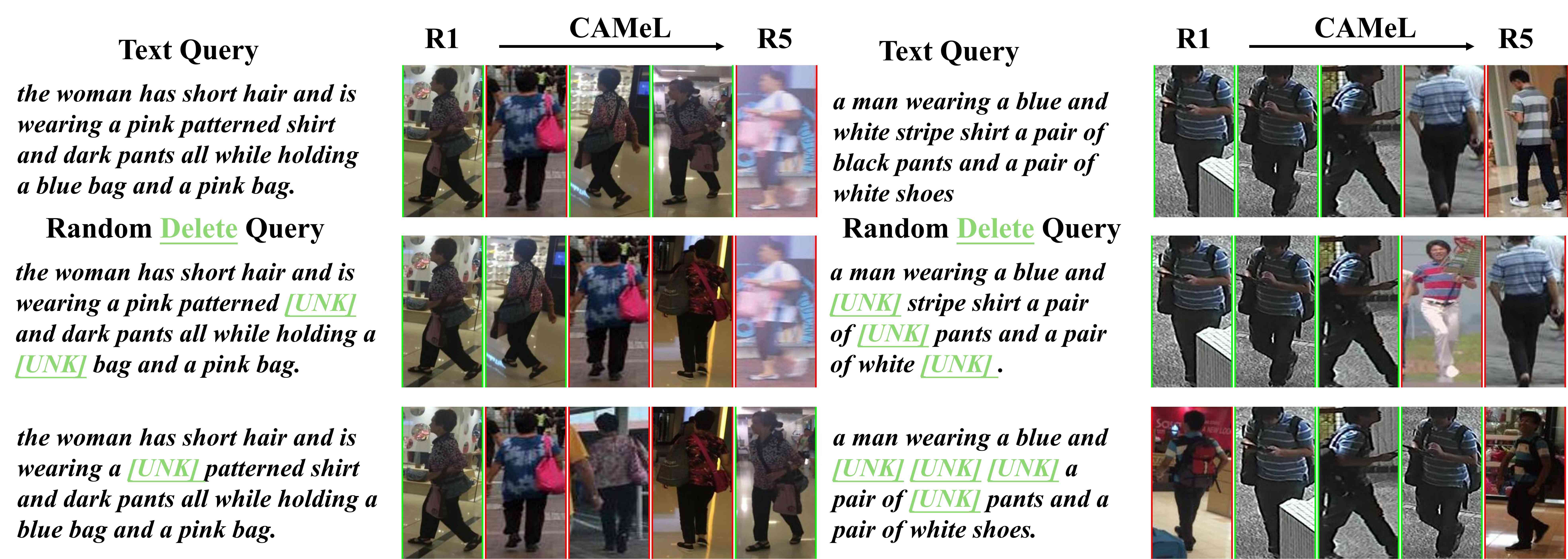}
    \caption{An example of person retrieval results based on text with randomly masking words is depicted. The retrieved images are arranged from left to right in descending order from R1 to R5. The results validate that increasing the number of deleted words does not impact the precision of our retrieval, confirming the robustness of the CAMeL. 
    The top image-text pair represents the original retrieval result. Green boxes indicate correct matches, while images in red boxes represent incorrect matches.}
    \vspace{-.1in}
    \label{fig8}
\end{figure}

\noindent\textbf{Effectiveness of Cross-Modality Meta-Learning.} 
To prevent model overfitting to specific types of tasks during training, which would impede effective learning of image features, we incorporate Cross-Modal Meta-Learning (CMML) as the core of our training strategy. As shown in Table~\ref{tab4}, we establish control groups using Stylization Tasks (ST) alone and compare them with experiments that include CMML. On the CUHK-PEDES dataset, we achieve improvements of 1.28\%, 0.72\%, 0.49\%, and 1.50\% on Recall@1, 5, 10, and mAP, respectively. Similar results are observed on the RSTPReid and ICFG-PEDES datasets, as shown in Table~\ref{tab4}. This highlights the role of CMML in facilitating better alignment between textual and visual data, which is crucial for fine-grained person retrieval. 
To ensure the fairness and robustness of the training pipeline, we further investigate the impact of pretraining and fine-tuning epochs on model performance. In our setting, the model is pretrained for 32 epochs on the synthetic dataset MALS to learn transferable representations, followed by 30 epochs of fine-tuning on each target dataset to balance generalization and domain-specific adaptation. To validate the effectiveness of this configuration, we conduct additional experiments by extending the fine-tuning epochs to 40, 50, and 60. Results show that prolonged fine-tuning does not lead to performance gains and instead introduces overfitting. Notably, on the CUHK-PEDES dataset, mAP and Recall@1 drop from 68.32\% to 66.35\% and from 77.24\% to 75.48\%, respectively. These findings confirm the rationality of the 32 epoch pretraining and 30 epoch fine-tuning configuration, which consistently yields optimal performance across datasets.

\begin{figure}[t]
    \centering
    \includegraphics[width=1\linewidth]{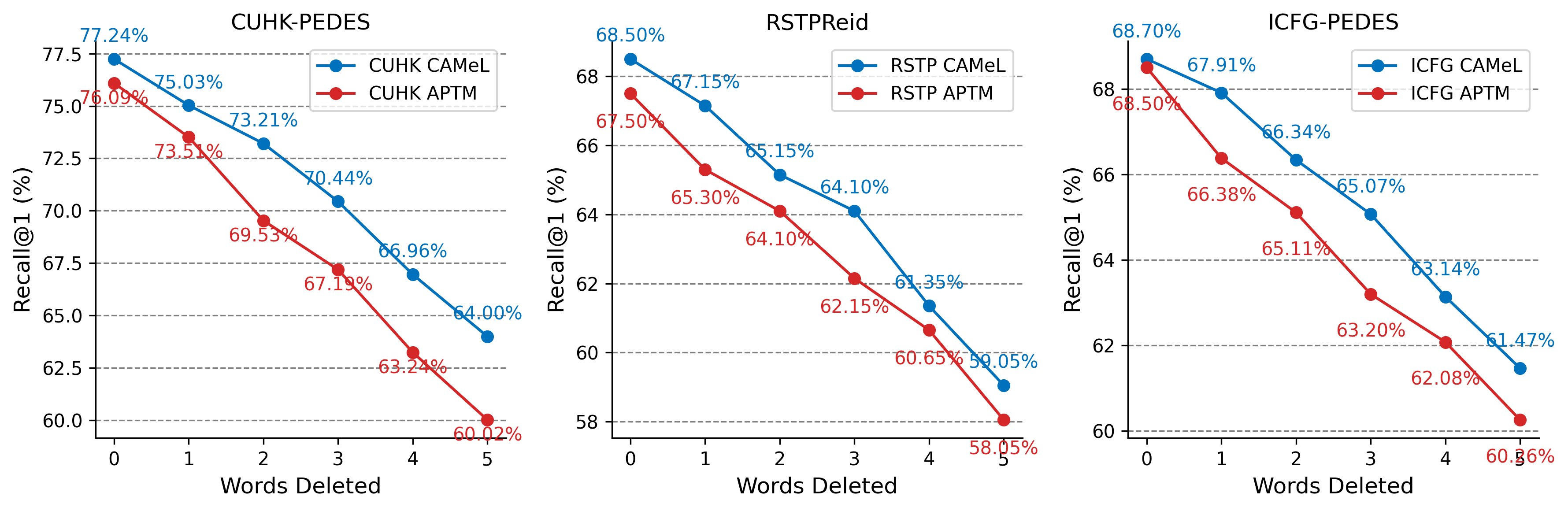}
    \caption{We assess model performance on the three benchmark datasets by randomly masking words from image annotations and comparing performance before and after deletion. The graph shows Ours (CAMeL) in blue and the APTM* in orange. We could observe that the proposed method is more robust against the ill-posed sentence queries (\eg, missing some words). 
    }
    \vspace{-.1in}
    \label{fig6}
\end{figure}

\noindent\textbf{Robustness against Ill-formed Text Query.} 
Our pre-trained model validates exceptional robustness and generalization capabilities in downstream text-based person retrieval tasks, even with incomplete text queries. After fine-tuning the CAMeL on the CUHK-PEDES dataset, we evaluate its performance by randomly masking 0 to 5 keywords in text queries with the special token [UNK]. As shown in Fig.~\ref{fig8}, despite the removal of crucial information such as ``clothing," ``color," and ``style," the model consistently maintains high retrieval accuracy. For example, the model accurately retrieves the image of a woman, even when the word ``color" is omitted from the description. Similarly, although the deletion of words like ``shirt" and ``pants" in the description of the man in the second row leads to one incorrect retrieval, the model still aligned well with the remaining text, illustrating its robustness. Furthermore, we compare our model against the APTM across three datasets using the same word masking technique. As shown in Fig.~\ref{fig6}, the gap in Recall@1 between our model and APTM increases significantly, from an initial difference of 1.15\% to 3.98\% on the CUHK-PEDES dataset, with our model also demonstrating more stable performance across the remaining datasets. This uniform enhancement confirms that our model exhibits significantly less fluctuation in Recall@1 with each additional word masking, proving its superior robustness compared to APTM across varied datasets.

\begin{figure}[t]
    \centering
    \includegraphics[width=0.7\linewidth]{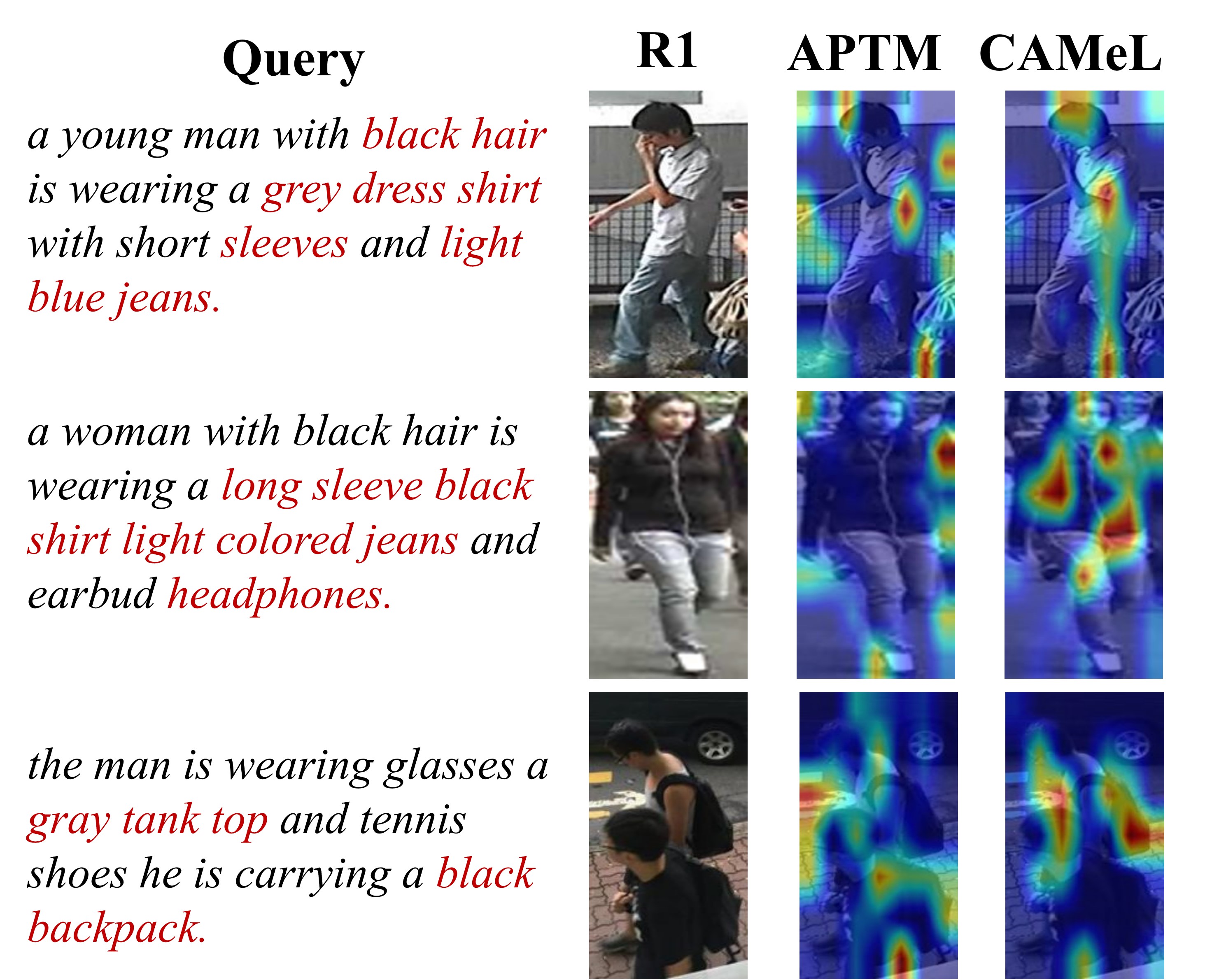}
    \caption{Visual comparison of cross attention maps generated by the APTM~\cite{yang2023towards11} and Ours (CAMeL) using Grad-CAM~\cite{jacobgilpytorchcam}. We could observe that the regions highlighted by the proposed methodology exhibit significant alignment with the keywords used in the query sentences, indicating effective matching performance.
    }
    \vspace{-.1in}
    \label{fig10}
\end{figure}

\noindent\textbf{Attention Comparison.} 
In Fig.~\ref{fig10}, several visual comparisons between the APTM and Ours are presented. Specifically, we utilize the Grad-CAM algorithm~\cite{jacobgilpytorchcam} to extract attention maps from the models, where each attention map shows the association between the query annotation and the retrieved full-body image of a person. It is evident that the model trained with our method has more focused and consistent attention on each attribute-related word, particularly on attributes such as ``grey dress shirt" and ``light blue jeans", where the attention accurately covers the corresponding objects. Furthermore, our training strategy allows the model to generate attention maps that more clearly focus on the correct attributes. For instance, in the ``headphones" and ``black backpack" attributes, our model exhibits more uniform and reasonable attention distribution compared to the baseline model. In contrast, the APTM model produces irrelevant attention noise in certain image regions, with more scattered attention distribution. Additionally, it is worth noting that even when there are multiple targets present in the image, our training strategy effectively avoids interference from other targets, ensuring that the model can accurately find and focus on the described target. These qualitative results further validate the effectiveness of our proposed training strategy in cross-modal tasks, as it enhances the precise association between text and images, which is crucial for text-based person retrieval tasks.

\begin{figure}[t]
    \centering
    \includegraphics[width=0.7\linewidth]{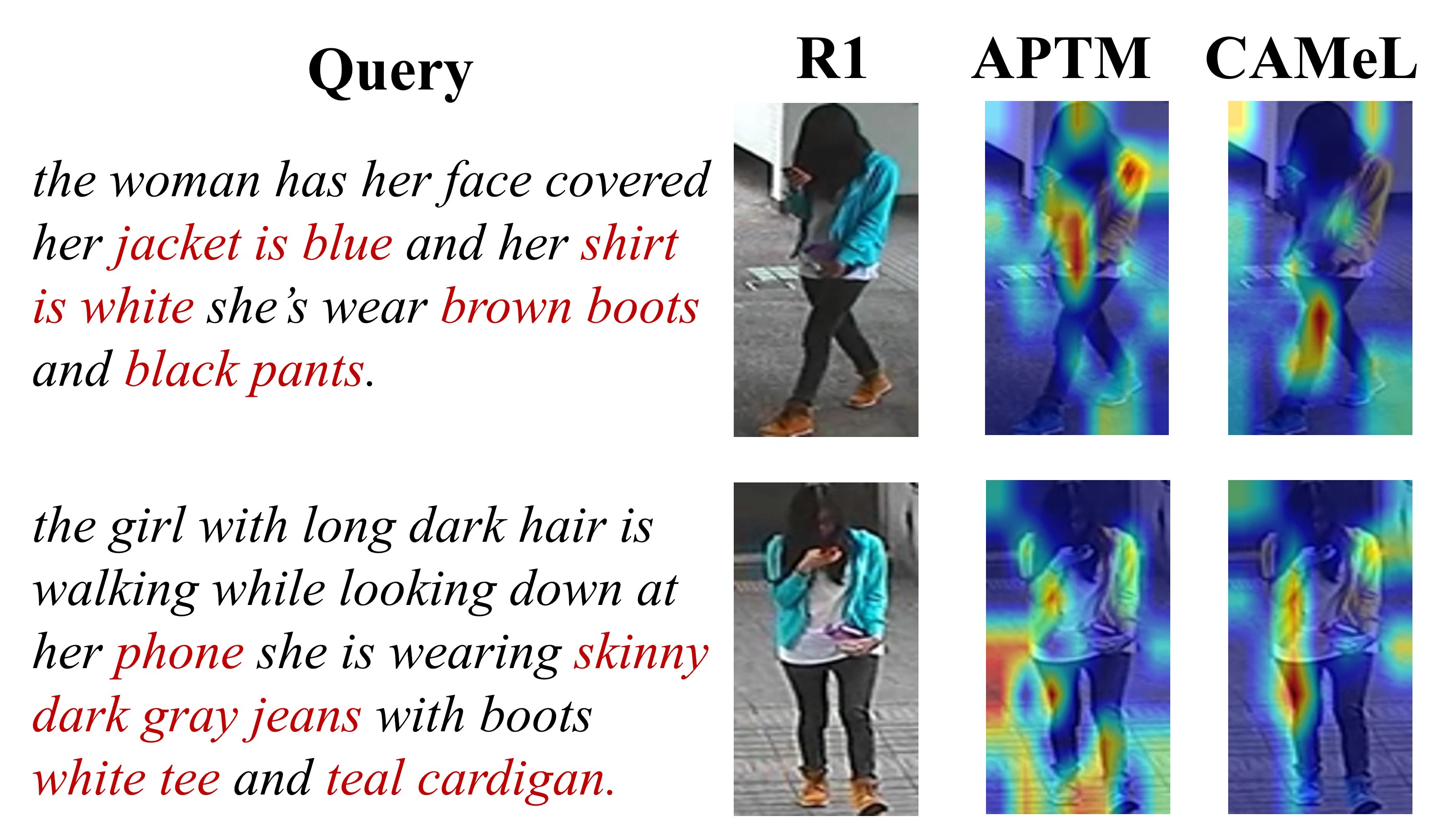}
    \caption{Comparative attention maps with varying queries for the same person.}
    \vspace{-.1in}
    \label{fig15}
\end{figure}

As shown in Fig~\ref{fig15}, the model has effectively captured the pedestrian and their clothing features, aligning well with the textual descriptions. Differences in viewpoint or descriptive emphasis lead to slight shifts in attention, occasionally focusing on background areas. Although the overall performance remains stable, there is still room for improvement—such as further enhancing multi-view robustness, refining text-image alignment, and suppressing irrelevant background in complex scenarios—to achieve more precise feature localization under higher precision and more diverse conditions.

\begin{table}[t]
\caption{\textbf{Comparison with other pretrained method.} We adopt CUHK-PEDES (denoted as C), ICFG-PEDES (denoted as I) and RSTPReid (denoted as R) as the source domain and the target domain in turn. R@K is Recall@K (higher is better). APTM*: We re-implement APTM~\cite{yang2023towards11}.}
\centering
\small
\sisetup{table-format=2.2} 
\begin{tabular}{
  l|
  S[table-format=2.2]
  S[table-format=2.2]
  S[table-format=2.2]
  S[table-format=2.2]
}
\toprule
 & {Method} & {R1} & {R5} & {R10}  \\
 \midrule	
 \multirow{7}{*}{C→I} 
 & {Dual Path~\cite{zheng2020dual8}} & 15.41 & 29.80 & 38.19 \\
 & {MIA~\cite{niu2020improving23}} & 19.35 & 36.78 & 46.42  \\
 & {SSAN~\cite{ding2021semantically12}} & 24.72 & 43.43 & 53.01  \\
 & {LGUR~\cite{shao2022learning32}} & 34.25 & 52.58 & 60.85  \\
 & {VGSG~\cite{he2023vgsg25}} & 35.85 & 55.04 & 63.61  \\
 & {APTM*~\cite{yang2023towards11}} & 48.57 & 67.06 & 74.02  \\
 \cline{2-5}
 & {Ours} & \textbf{49.18} & \textbf{67.58} & \textbf{74.63}  \\
\midrule	
 \multirow{7}{*}{I→C} 
& {Dual Path~\cite{zheng2020dual8}} & 7.63 & 17.14 & 23.52 \\
 & {MIA~\cite{niu2020improving23}} & 10.93 & 23.77 & 32.39  \\
 & {SSAN~\cite{ding2021semantically12}} & 16.68 & 33.84 & 43.00  \\
 & {LGUR~\cite{shao2022learning32}} & 25.44 & 44.48 & 54.39  \\
 & {VGSG~\cite{he2023vgsg25}} & 27.17 & 47.77 & 57.27  \\
 & {APTM*~\cite{yang2023towards11}} & 46.52 & 67.53 & 76.27 \\
  \cline{2-5}
 & {Ours} & \textbf{70.66} & \textbf{87.09} & \textbf{91.91}  \\
 \midrule
  \multirow{2}{*}{I→R} 
 & {APTM*} & 54.75 & 77.45 & 83.90  \\
 & {Ours} & \textbf{60.15} & \textbf{80.15} & \textbf{87.15}  \\
\midrule	
 \multirow{2}{*}{R→I} 
 & {APTM*} & 43.11 & 58.79 & 65.89 \\
 & {Ours} & \textbf{46.34} & \textbf{62.88} & \textbf{69.81}  \\
 
\bottomrule
\end{tabular}
\vspace{-.1in}
\label{tab6}
\end{table}

\noindent\textbf{Zero-shot Learning.} 
To further validate the generalization capability of the CAMeL, we perform zero-shot experiments on three datasets: CUHK-PEDES~\cite{li2017person1}, RSTPReid~\cite{zhu2021dssl13} and ICFG-PEDES\cite{ding2021semantically12}. Without any fine-tuning, we directly test the pre-trained model on the target dataset under identical parameter settings. As shown in Table~\ref{tab1},~\ref{tab2}, and~\ref{tab3}, our model demonstrates superior generalization ability, achieving improvements in Recall@1, Recall@5, and Recall@10. These results underscore the robustness of the CAMeL strategy employed during pretraining, enabling the model to perform well on unseen datasets without the need for domain-specific fine-tuning. This suggests that the CAMeL approach effectively enhances the model's initial performance, improving its capacity to generalize across different tasks and domains.

\noindent\textbf{Domain Migration.} 
We also conduct domain adaptation experiments to further evaluate the robustness of model. Specifically, the CAMeL fine-tuned on CUHK-PEDES is applied to the ICFG-PEDES dataset. Since both ICFG-PEDES and RSTPReid are derived from the same  
pretraining dataset, focusing on the CUHK-PEDES to ICFG-PEDES migration ensures a meaningful and challenging domain shift, while also avoiding redundancy from testing two closely related datasets. As shown in Table~\ref{tab6}, the model demonstrates stable performance on ICFG-PEDES, validating its ability to adapt to new but related domains.
Additionally, we perform a reverse domain migration by fine-tuning the model on RSTPReid and testing it on CUHK-PEDES. This experiment allows us to examine the generalization of model capability when moving from a more controlled surveillance dataset to one with more diverse and varied visual characteristics. The results, as shown in Table~\ref{tab6}, reveal that our model maintains competitive performance, illustrating its robustness in adapting to cross-domain variations without extensive re-training.
Lastly, we conduct migration experiments between ICFG-PEDES and RSTPReid. Testing these migrations provides additional insights into the model’s fine-grained domain adaptation capabilities. The results further support the generalization of model strength across datasets that are both related and distinct in their own ways.

\section{Conclusion}
\label{Sec.5}
In this paper, we introduce a domain-agnostic pretraining framework that integrates stylized tasks, cross-modality meta-learning, and an adaptive dual-speed strategy to mitigate the negative impact of bias in generated data on model generalization capabilities. By designing stylized tasks, our data better simulates the diversity and complexity of the real world. With the aid of cross-modal meta-learning, we achieve effective integration of information across different tasks, thereby enhancing the generalizability of model. Furthermore, the adaptive dual-speed update strategy delves into the specifics of each task, allowing the model to fast absorb new knowledge while meticulously optimizing the long-term learning process.
Experimental results verify that our approach achieves competitive recall rates on the three benchmark datasets. This validates the effectiveness of our proposed pretraining strategy in complex environments. 
In the future, we plan to explore mechanisms based on erroneous samples to enhance model accuracy. We will probe deeper into the feature representation of erroneous samples. By dynamically identifying and reintegrating these samples, we aim to improve adaptability in challenging scenarios.

\section{Acknowledgment}
This work was supported by projects of the Shanghai Committee of Science and Technology, China (No.23ZR1423500), the National Natural Science Foundation of China (No.62302287), the Nanjing Municipal Science and Technology Bureau (No.202401035), the Guangdong Basic and Applied Basic Research Foundation (No.2025A1515012281).

\bibliography{reference}

\begin{thebibliography}{10}
\expandafter\ifx\csname url\endcsname\relax
  \def\url#1{\texttt{#1}}\fi
\expandafter\ifx\csname urlprefix\endcsname\relax\def\urlprefix{URL }\fi
\expandafter\ifx\csname href\endcsname\relax
  \def\href#1#2{#2} \def\path#1{#1}\fi

\bibitem{book}
Z.~Zheng, L.~Zheng, Object re-identification: Problems, algorithms and responsible research practice, in: The Boundaries of Data, Amsterdam University Press, 2024.

\bibitem{xiao2016end4}
T.~Xiao, S.~Li, B.~Wang, L.~Lin, X.~Wang, End-to-end deep learning for person search, arXiv preprint arXiv:1604.01850 2~(2) (2016) 4.

\bibitem{jiang2023cross6}
D.~Jiang, M.~Ye, Cross-modal implicit relation reasoning and aligning for text-to-image person retrieval, in: Proceedings of the IEEE/CVF Conference on Computer Vision and Pattern Recognition (CVPR), 2023, pp. 2787--2797.

\bibitem{10.1145/3543507.3583254}
H.~Shan, Q.~Zhang, Z.~Liu, G.~Zhang, C.~Li, Beyond two-tower: Attribute guided representation learning for candidate retrieval, in: Proceedings of the ACM Web Conference 2023 (WWW), 2023, pp. 3173--3181.

\bibitem{li2017person1}
S.~Li, T.~Xiao, H.~Li, B.~Zhou, D.~Yue, X.~Wang, Person search with natural language description, in: Proceedings of the IEEE Conference on Computer Vision and Pattern Recognition (CVPR), 2017, pp. 1970--1979.

\bibitem{nguyen2024ag}
H.~Nguyen, K.~Nguyen, S.~Sridharan, C.~Fookes, Ag-reid. v2: Bridging aerial and ground views for person re-identification, IEEE Transactions on Information Forensics and Security (TIFS)Doi:10.1109/TIFS.2024.3353078 (2024).

\bibitem{cheng2024neighbor101}
D.~Cheng, H.~Tai, N.~Wang, C.~Fang, X.~Gao, Neighbor consistency and global-local interaction: A novel pseudo-label refinement approach for unsupervised person re-identification, IEEE Transactions on Information Forensics and Security (TIFS)Doi:10.1109/TIFS.2024.3465037 (2024).

\bibitem{lu2025nighttime100}
A.~Lu, C.~Li, T.~Zha, X.~Wang, J.~Tang, B.~Luo, Nighttime person re-identification via collaborative enhancement network with multi-domain learning, IEEE Transactions on Information Forensics and Security (TIFS)Doi:10.1109/TIFS.2025.3527335 (2025).

\bibitem{liu2024improving102}
Y.~Liu, M.~Qi, Y.~Zhang, Q.~Wu, J.~Wu, S.~Zhuang, Improving consistency of proxy-level contrastive learning for unsupervised person re-identification, IEEE Transactions on Information Forensics and Security (TIFS)Doi:10.1109/TIFS.2024.3426351 (2024).

\bibitem{ye2024securereid103}
M.~Ye, W.~Shen, J.~Zhang, Y.~Yang, B.~Du, Securereid: Privacy-preserving anonymization for person re-identification, IEEE Transactions on Information Forensics and Security (TIFS)Doi:10.1109/TIFS.2024.3356233 (2024).

\bibitem{li2018richly64}
D.~Li, Z.~Zhang, X.~Chen, K.~Huang, A richly annotated pedestrian dataset for person retrieval in real surveillance scenarios, IEEE Transactions on Image Processing (TIP) 28~(4) (2018) 1575--1590, doi:10.1109/TIP.2018.2878349.

\bibitem{zheng2017unlabeled}
Z.~Zheng, L.~Zheng, Y.~Yang, Unlabeled samples generated by gan improve the person re-identification baseline in vitro, in: Proceedings of the IEEE International Conference on Computer Vision (ICCV), 2017, pp. 3754--3762.

\bibitem{bertocco2021unsupervised}
G.~C. Bertocco, F.~Andaló, A.~Rocha, Unsupervised and self-adaptative techniques for cross-domain person re-identification, IEEE Transactions on Information Forensics and Security (TIFS) 16 (2021) 4419--4434, doi:10.1109/TIFS.2021.3107157.

\bibitem{yang2023towards11}
S.~Yang, Y.~Zhou, Z.~Zheng, Y.~Wang, L.~Zhu, Y.~Wu, Towards unified text-based person retrieval: A large-scale multi-attribute and language search benchmark, in: Proceedings of the 31st ACM International Conference on Multimedia (ACMM), 2023, pp. 4492--4501.

\bibitem{chu2024towards}
M.~Chu, Z.~Zheng, W.~Ji, T.~Wang, T.-S. Chua, Towards natural language-guided drones: Geotext-1652 benchmark with spatial relation matching, in: Proceedings of the European Conference on Computer Vision (ECCV), 2024.

\bibitem{yao2023capenrich}
L.~Yao, W.~Chen, Q.~Jin, Capenrich: Enriching caption semantics for web images via cross-modal pre-trained knowledge, in: Proceedings of the ACM Web Conference 2023 (WWW), 2023, pp. 2392--2401.

\bibitem{rombach2022high9}
R.~Rombach, A.~Blattmann, D.~Lorenz, P.~Esser, B.~Ommer, High-resolution image synthesis with latent diffusion models, in: Proceedings of the IEEE/CVF Conference on Computer Vision and Pattern Recognition (CVPR), 2022, pp. 10684--10695.

\bibitem{hertz2022prompt10}
A.~Hertz, R.~Mokady, J.~Tenenbaum, K.~Aberman, Y.~Pritch, D.~Cohen-Or, Prompt-to-prompt image editing with cross attention control, arXiv preprint arXiv:2208.01626 (2022).

\bibitem{karras2017progressive}
T.~Karras, Progressive growing of gans for improved quality, stability, and variation, arXiv preprint arXiv:1710.10196 (2017).

\bibitem{ding2021semantically12}
Z.~Ding, C.~Ding, Z.~Shao, D.~Tao, Semantically self-aligned network for text-to-image part-aware person re-identification, arXiv preprint arXiv:2107.12666 (2021).

\bibitem{zhu2021dssl13}
A.~Zhu, Z.~Wang, Y.~Li, X.~Wan, J.~Jin, T.~Wang, F.~Hu, G.~Hua, Dssl: Deep surroundings-person separation learning for text-based person retrieval, in: Proceedings of the 29th ACM International Conference on Multimedia (ACMM), 2021, pp. 209--217.

\bibitem{shao2022learning32}
Z.~Shao, X.~Zhang, M.~Fang, Z.~Lin, J.~Wang, C.~Ding, Learning granularity-unified representations for text-to-image person re-identification, in: Proceedings of the 30th ACM International Conference on Multimedia (ACMM), 2022, pp. 5566--5574.

\bibitem{bai2023rasa33}
Y.~Bai, M.~Cao, D.~Gao, Z.~Cao, C.~Chen, Z.~Fan, L.~Nie, M.~Zhang, Rasa: Relation and sensitivity aware representation learning for text-based person search, arXiv preprint arXiv:2305.13653 (2023).

\bibitem{fujii2023bilma34}
T.~Fujii, S.~Tarashima, Bilma: Bidirectional local-matching for text-based person re-identification, in: Proceedings of the IEEE/CVF International Conference on Computer Vision (ICCV), 2023, pp. 2786--2790.

\bibitem{wang2021dynamic21}
S.~Wang, H.~Li, Z.~Wang, W.~Ouyang, Dynamic position-aware network for fine-grained image recognition, in: Proceedings of the AAAI Conference on Artificial Intelligence (AAAI), Vol.~35, 2021, pp. 2791--2799.

\bibitem{wang2023progressive}
X.~Wang, Z.~Zheng, Y.~He, F.~Yan, Z.~Zeng, Y.~Yang, Progressive local filter pruning for image retrieval acceleration, IEEE Transactions on Multimedia 25 (2023) 9597--9607, doi:10.1109/TMM.2023.3256092.

\bibitem{niu2020improving23}
K.~Niu, Y.~Huang, W.~Ouyang, L.~Wang, Improving description-based person re-identification by multi-granularity image-text alignments, IEEE Transactions on Image Processing (TIP) 29 (2020) 5542--5556, doi:10.1109/TIP.2020.2984883.

\bibitem{sarafianos2019adversarial24}
N.~Sarafianos, X.~Xu, I.~A. Kakadiaris, Adversarial representation learning for text-to-image matching, in: Proceedings of the IEEE/CVF International Conference on Computer Vision (ICCV), 2019, pp. 5814--5824.

\bibitem{he2023vgsg25}
S.~He, H.~Luo, W.~Jiang, X.~Jiang, H.~Ding, Vgsg: Vision-guided semantic-group network for text-based person search, IEEE Transactions on Image Processing (TIP) 33 (2023) 163--176, doi:10.1109/TIP.2023.3337653.

\bibitem{ergasti2024mars}
A.~Ergasti, T.~Fontanini, C.~Ferrari, M.~Bertozzi, A.~Prati, Mars: Paying more attention to visual attributes for text-based person search, arXiv preprint arXiv:2407.04287 (2024).

\bibitem{tan2024occluded}
L.~Tan, J.~Xia, W.~Liu, P.~Dai, Y.~Wu, L.~Cao, Occluded person re-identification via saliency-guided patch transfer, in: Proceedings of the AAAI Conference on Artificial Intelligence (AAAI), Vol.~38, 2024, pp. 5070--5078.

\bibitem{zhang2024adaptive}
Y.~Zhang, Y.~Yan, Y.~Lu, H.~Wang, Adaptive middle modality alignment learning for visible-infrared person re-identification, International Journal of Computer Vision (IJCV) (2024) 1--21Doi:10.1007/s11263-024-02276-4.

\bibitem{tan2024rle}
L.~Tan, Y.~Zhang, K.~Han, P.~Dai, Y.~Zhang, Y.~Wu, R.~Ji, Rle: A unified perspective of data augmentation for cross-spectral re-identification, arXiv preprint arXiv:2411.01225 (2024).

\bibitem{verma2021towards66}
V.~Verma, T.~Luong, K.~Kawaguchi, H.~Pham, Q.~Le, Towards domain-agnostic contrastive learning, in: International Conference on Machine Learning (ICML), PMLR, 2021, pp. 10530--10541.

\bibitem{tamkin2021dabs67}
A.~Tamkin, V.~Liu, R.~Lu, D.~Fein, C.~Schultz, N.~Goodman, Dabs: A domain-agnostic benchmark for self-supervised learning, arXiv preprint arXiv:2111.12062 (2021).

\bibitem{lee2020mix68}
K.~Lee, Y.~Zhu, K.~Sohn, C.-L. Li, J.~Shin, H.~Lee, I-mix: A domain-agnostic strategy for contrastive representation learning, arXiv preprint arXiv:2010.08887 (2020).

\bibitem{mishra2022task2sim70}
S.~Mishra, R.~Panda, C.~P. Phoo, C.-F.~R. Chen, L.~Karlinsky, K.~Saenko, V.~Saligrama, R.~S. Feris, Task2sim: Towards effective pre-training and transfer from synthetic data, in: Proceedings of the IEEE/CVF Conference on Computer Vision and Pattern Recognition (CVPR), 2022, pp. 9194--9204.

\bibitem{zhu2024mario}
Y.~Zhu, H.~Shi, Z.~Zhang, S.~Tang, Mario: Model agnostic recipe for improving ood generalization of graph contrastive learning, in: Proceedings of the ACM on Web Conference 2024 (WWW), 2024, pp. 300--311.

\bibitem{huo2022domain71}
X.~Huo, L.~Xie, H.~Hu, W.~Zhou, H.~Li, Q.~Tian, Domain-agnostic prior for transfer semantic segmentation, in: Proceedings of the IEEE/CVF Conference on Computer Vision and Pattern Recognition (CVPR), 2022, pp. 7075--7085.

\bibitem{du2024domain72}
Z.~Du, X.~Li, F.~Li, K.~Lu, L.~Zhu, J.~Li, Domain-agnostic mutual prompting for unsupervised domain adaptation, arXiv preprint arXiv:2403.02899 (2024).

\bibitem{hu2023large}
J.~Hu, Y.~Yao, C.~Wang, S.~Wang, Y.~Pan, Q.~Chen, T.~Yu, H.~Wu, Y.~Zhao, H.~Zhang, et~al., Large multilingual models pivot zero-shot multimodal learning across languages, arXiv preprint arXiv:2308.12038 (2023).

\bibitem{lv2024rethinking}
F.~Lv, C.~Nie, J.~Zhang, G.~Yang, G.~Lin, X.~Wu, T.~Li, Rethinking the effect of uninformative class name in prompt learning, in: Proceedings of the 32nd ACM International Conference on Multimedia (ACMM), 2024, pp. 8345--8354.

\bibitem{finn2017model35}
C.~Finn, P.~Abbeel, S.~Levine, Model-agnostic meta-learning for fast adaptation of deep networks, in: International Conference on Machine Learning (ICML), PMLR, 2017, pp. 1126--1135.

\bibitem{nichol2018reptile31}
A.~Nichol, J.~Schulman, Reptile: A scalable metalearning algorithm, arXiv preprint arXiv:1803.02999 2~(3) (2018) 4.

\bibitem{10.1145/3589334.364536990}
Z.~Zhang, Q.~Liu, Z.~Hu, Y.~Zhan, Z.~Huang, W.~Gao, Q.~Mao, Enhancing fairness in meta-learned user modeling via adaptive sampling, in: Proceedings of the ACM Web Conference 2024 (WWW), 2024, pp. 3241--3252.

\bibitem{li2024adaptive83}
S.~Li, C.~He, X.~Xu, F.~Shen, Y.~Yang, H.~T. Shen, Adaptive uncertainty-based learning for text-based person retrieval, in: Proceedings of the AAAI Conference on Artificial Intelligence (AAAI), Vol.~38, 2024, pp. 3172--3180.

\bibitem{ma2022multimodality84}
Y.~Ma, S.~Zhao, W.~Wang, Y.~Li, I.~King, Multimodality in meta-learning: A comprehensive survey, Knowledge-Based Systems 250 (2022) 108976, doi:10.1016/j.knosys.2022.108976.

\bibitem{10.1145/3543507.3583548}
Z.~Tian, Z.~Xie, F.~Lin, Y.~Song, A multi-view meta-learning approach for multi-modal response generation, in: Proceedings of the ACM Web Conference 2023 (WWW), 2023, pp. 1938--1947.

\bibitem{10.1145/3485447.3512013}
X.~Wang, L.~Cao, H.~Zhang, L.~Feng, Y.~Ding, N.~Li, A meta-learning based stress category detection framework on social media, in: Proceedings of the ACM Web Conference 2022 (WWW), 2022, pp. 2925--2935.

\bibitem{sun2019meta85}
Q.~Sun, Y.~Liu, T.-S. Chua, B.~Schiele, Meta-transfer learning for few-shot learning, in: Proceedings of the IEEE/CVF Conference on Computer Vision and Pattern Recognition (CVPR), 2019, pp. 403--412.

\bibitem{tran20233fm86}
M.~Tran, R.~Shah, Z.~Gong, 3fm: Multi-modal meta-learning for federated tasks, arXiv preprint arXiv:2312.10179 (2023).

\bibitem{wang2022look57}
Z.~Wang, A.~Zhu, J.~Xue, X.~Wan, C.~Liu, T.~Wang, Y.~Li, Look before you leap: Improving text-based person retrieval by learning a consistent cross-modal common manifold, in: Proceedings of the 30th ACM International Conference on Multimedia (ACMM), 2022, pp. 1984--1992.

\bibitem{vettoruzzo2024advances88}
A.~Vettoruzzo, M.-R. Bouguelia, J.~Vanschoren, T.~Rognvaldsson, K.~Santosh, Advances and challenges in meta-learning: A technical review, IEEE Transactions on Pattern Analysis and Machine Intelligence (TPAMI)Doi:10.1109/TPAMI.2024.3357847 (2024).

\bibitem{Wei_Zou_201927}
J.~Wei, K.~Zou, Eda: Easy data augmentation techniques for boosting performance on text classification tasks, in: Proceedings of the 2019 Conference on Empirical Methods in Natural Language Processing and the 9th International Joint Conference on Natural Language Processing (EMNLP-IJCNLP), 2019.

\bibitem{Cubuk_Zoph_Shlens_Le_202028}
E.~D. Cubuk, B.~Zoph, J.~Shlens, Q.~V. Le, Randaugment: Practical automated data augmentation with a reduced search space, in: 2020 IEEE/CVF Conference on Computer Vision and Pattern Recognition Workshops (CVPRW), 2020.

\bibitem{han2022g30}
X.~Han, Z.~Jiang, N.~Liu, X.~Hu, G-mixup: Graph data augmentation for graph classification, in: International Conference on Machine Learning (ICML), PMLR, 2022, pp. 8230--8248.

\bibitem{zhong2024memorybank41}
W.~Zhong, L.~Guo, Q.~Gao, H.~Ye, Y.~Wang, Memorybank: Enhancing large language models with long-term memory, in: Proceedings of the AAAI Conference on Artificial Intelligence (AAAI), Vol.~38, 2024, pp. 19724--19731.

\bibitem{wang2022point73}
Z.~Wang, Z.~Gao, X.~Xu, Y.~Luo, Y.~Yang, H.~T. Shen, Point to rectangle matching for image text retrieval, in: Proceedings of the 30th ACM International Conference on Multimedia (ACMM), 2022, pp. 4977--4986.

\bibitem{nichol2018first76}
A.~Nichol, J.~Achiam, J.~Schulman, On first-order meta-learning algorithms, arXiv preprint arXiv:1803.02999 (2018).

\bibitem{ravi2016optimization77}
S.~Ravi, H.~Larochelle, Optimization as a model for few-shot learning, in: International Conference on Learning Representations (ICLR), 2016.

\bibitem{wu2019ace78}
Z.~Wu, X.~Wang, J.~E. Gonzalez, T.~Goldstein, L.~S. Davis, Ace: Adapting to changing environments for semantic segmentation, in: Proceedings of the IEEE/CVF International Conference on Computer Vision (ICCV), 2019, pp. 2121--2130.

\bibitem{robbins1951stochastic82}
H.~Robbins, S.~Monro, A stochastic approximation method, The Annals of Mathematical Statistics (1951) 400--407.

\bibitem{wei2018person42}
L.~Wei, S.~Zhang, W.~Gao, Q.~Tian, Person transfer gan to bridge domain gap for person re-identification, in: Proceedings of the IEEE Conference on Computer Vision and Pattern Recognition (CVPR), 2018, pp. 79--88.

\bibitem{zheng2020dual8}
Z.~Zheng, L.~Zheng, M.~Garrett, Y.~Yang, M.~Xu, Y.-D. Shen, Dual-path convolutional image-text embeddings with instance loss, ACM Transactions on Multimedia Computing, Communications, and Applications (TOMM) 16~(2) (2020) 1--23, doi:10.1145/3383184.

\bibitem{zhang2018deep7}
Y.~Zhang, H.~Lu, Deep cross-modal projection learning for image-text matching, in: Proceedings of the European Conference on Computer Vision (ECCV), 2018, pp. 686--701.

\bibitem{liu2019deep48}
J.~Liu, Z.-J. Zha, R.~Hong, M.~Wang, Y.~Zhang, Deep adversarial graph attention convolution network for text-based person search, in: Proceedings of the 27th ACM International Conference on Multimedia (ACMM), 2019, pp. 665--673.

\bibitem{wang2020vitaa49}
Z.~Wang, Z.~Fang, J.~Wang, Y.~Yang, Vitaa: Visual-textual attributes alignment in person search by natural language, in: Proceedings of the European Conference on Computer Vision (ECCV), Springer, 2020, pp. 402--420.

\bibitem{wang2020img50}
Z.~Wang, A.~Zhu, Z.~Zheng, J.~Jin, Z.~Xue, G.~Hua, Img-net: Inner-cross-modal attentional multigranular network for description-based person re-identification, Journal of Electronic Imaging 29~(4) (2020) 043028--043028, doi:10.1117/1.JEI.29.4.043028.

\bibitem{aggarwal2020text51}
S.~Aggarwal, V.~B. Radhakrishnan, A.~Chakraborty, Text-based person search via attribute-aided matching, in: Proceedings of the IEEE/CVF Winter Conference on Applications of Computer Vision (WACV), 2020, pp. 2617--2625.

\bibitem{zheng2020hierarchical52}
K.~Zheng, W.~Liu, J.~Liu, Z.-J. Zha, T.~Mei, Hierarchical gumbel attention network for text-based person search, in: Proceedings of the 28th ACM International Conference on Multimedia (ACMM), 2020, pp. 3441--3449.

\bibitem{gao2021contextual53}
C.~Gao, G.~Cai, X.~Jiang, F.~Zheng, J.~Zhang, Y.~Gong, P.~Peng, X.~Guo, X.~Sun, Contextual non-local alignment over full-scale representation for text-based person search, arXiv preprint arXiv:2101.03036 (2021).

\bibitem{wang2021text54}
C.~Wang, Z.~Luo, Y.~Lin, S.~Li, Text-based person search via multi-granularity embedding learning, in: International Joint Conference on Artificial Intelligence (IJCAI), 2021, pp. 1068--1074.

\bibitem{han2021text55}
X.~Han, S.~He, L.~Zhang, T.~Xiang, Text-based person search with limited data, arXiv preprint arXiv:2110.10807 (2021).

\bibitem{chen2022tipcb56}
Y.~Chen, G.~Zhang, Y.~Lu, Z.~Wang, Y.~Zheng, Tipcb: A simple but effective part-based convolutional baseline for text-based person search, Neurocomputing 494 (2022) 171--181, doi:10.1016/j.neucom.2022.04.081.

\bibitem{wang2022caibc58}
Z.~Wang, A.~Zhu, J.~Xue, X.~Wan, C.~Liu, T.~Wang, Y.~Li, Caibc: Capturing all-round information beyond color for text-based person retrieval, in: Proceedings of the 30th ACM International Conference on Multimedia (ACMM), 2022, pp. 5314--5322.

\bibitem{farooq2022axm59}
A.~Farooq, M.~Awais, J.~Kittler, S.~S. Khalid, Axm-net: Implicit cross-modal feature alignment for person re-identification, in: Proceedings of the AAAI Conference on Artificial Intelligence (AAAI), Vol.~36, 2022, pp. 4477--4485.

\bibitem{suo2022simple60}
W.~Suo, M.~Sun, K.~Niu, Y.~Gao, P.~Wang, Y.~Zhang, Q.~Wu, A simple and robust correlation filtering method for text-based person search, in: Proceedings of the European Conference on Computer Vision (ECCV), Springer, 2022, pp. 726--742.

\bibitem{yan2023clip61}
S.~Yan, N.~Dong, L.~Zhang, J.~Tang, Clip-driven fine-grained text-image person re-identification, IEEE Transactions on Image Processing (TIP)Doi:10.1109/TIP.2023.3327924 (2023).

\bibitem{zuo2023plip62}
J.~Zuo, C.~Yu, N.~Sang, C.~Gao, Plip: Language-image pre-training for person representation learning, arXiv preprint arXiv:2305.08386 (2023).

\bibitem{cao2024empirical63}
M.~Cao, Y.~Bai, Z.~Zeng, M.~Ye, M.~Zhang, An empirical study of clip for text-based person search, in: Proceedings of the AAAI Conference on Artificial Intelligence (AAAI), Vol.~38, 2024, pp. 465--473.

\bibitem{qin2023noisy}
Y.~Qin, Y.~Chen, D.~Peng, X.~Peng, J.~T. Zhou, P.~Hu, Noisy-correspondence learning for text-to-image person re-identification, arXiv preprint arXiv:2308.09911 (2023).

\bibitem{sun2024data}
J.~Sun, Z.~Zheng, G.~Ding, From data deluge to data curation: A filtering-wora paradigm for efficient text-based person search, Proceedings of the ACM on Web Conference (WWW) (2025).

\bibitem{ren2023sg38}
S.~Ren, X.~Yang, S.~Liu, X.~Wang, Sg-former: Self-guided transformer with evolving token reallocation, in: Proceedings of the IEEE/CVF International Conference on Computer Vision (ICCV), 2023, pp. 6003--6014.

\bibitem{devlin2018bert18}
J.~Devlin, M.-W. Chang, K.~Lee, K.~Toutanova, Bert: Pre-training of deep bidirectional transformers for language understanding, arXiv preprint arXiv:1810.04805 (2018).

\bibitem{loshchilov2017decoupled43}
I.~Loshchilov, F.~Hutter, Decoupled weight decay regularization, arXiv preprint arXiv:1711.05101 (2017).

\bibitem{izmailov2018averaging44}
P.~Izmailov, D.~Podoprikhin, T.~Garipov, D.~Vetrov, A.~G. Wilson, Averaging weights leads to wider optima and better generalization, arXiv preprint arXiv:1803.05407 (2018).

\bibitem{shu2022see40}
X.~Shu, W.~Wen, H.~Wu, K.~Chen, Y.~Song, R.~Qiao, B.~Ren, X.~Wang, See finer, see more: Implicit modality alignment for text-based person retrieval, in: Proceedings of the European Conference on Computer Vision (ECCV), Springer, 2022, pp. 624--641.

\bibitem{Zheng_Wei_Yang_202001}
Z.~Zheng, Y.~Wei, Y.~Yang, University-1652: A multi-view multi-source benchmark for drone-based geo-localization, in: Proceedings of the 28th ACM International Conference on Multimedia (ACMM), 2020, pp. 1395--1403.

\bibitem{zhu2023sues18}
R.~Zhu, L.~Yin, M.~Yang, F.~Wu, Y.~Yang, W.~Hu, Sues-200: A multi-height multi-scene cross-view image benchmark across drone and satellite, IEEE Transactions on Circuits and Systems for Video Technology (TCSVT) 33~(9) (2023) 4825--4839, doi:10.1109/TCSVT.2023.3249204.

\bibitem{Zhai_Bessinger_Workman_Jacobs_201717}
M.~Zhai, Z.~Bessinger, S.~Workman, N.~Jacobs, Predicting ground-level scene layout from aerial imagery, in: 2017 IEEE Conference on Computer Vision and Pattern Recognition (CVPR), 2017.

\bibitem{jacobgilpytorchcam}
J.~Gildenblat, contributors, Pytorch library for cam methods, \url{https://github.com/jacobgil/pytorch-grad-cam} (2021).

\end{thebibliography}
\bibliographystyle{elsarticle-num}

\end{document}